\documentclass[runningheads]{llncs}

\usepackage{graphicx}
\usepackage{comment}
\usepackage{amsmath,amssymb}
\usepackage{color}
\usepackage{url}
\usepackage{hyperref}
\usepackage{multirow}
\usepackage{algorithm}
\usepackage{algorithmic}
\usepackage{xcolor}
\usepackage{booktabs}
\usepackage{subcaption} 
\usepackage{pifont}
\usepackage{breakcites}
\usepackage{caption}
\usepackage{wrapfig}
\usepackage{adjustbox}

\newcommand{\mathbbm}[1]{\text{\usefont{U}{bbm}{m}{n}#1}}
\newcommand{\myparagraph}[1]{\vspace{2.3pt}\noindent{\bf #1}}
\newif\ifreview
\reviewtrue
\reviewfalse

\ifreview
	\usepackage{lineno}

	\linenumbers
\fi

\begin{document}

%%%%%%%%%%%%%%%%%%%%% Add submission id, track, and title. %%%%%%%%%%%%%%%%%%%%%

% Insert your submission number here
\def\SubNumber{060}

% Choose one track by uncommenting one of the following lines  
\def\GCPRTrack{Regular Track}
% \def\GCPRTrack{Track: Computer vision systems and applications}
% \def\GCPRTrack{Track: Pattern recognition in the life and natural sciences}
% \def\GCPRTrack{Track: Photogrammetry and remote sensing}
% \def\GCPRTrack{Track: Robot vision}
% \def\GCPRTrack{Track: DAGM Young Researcher Forum}

% Replace with your title
\title{Revisiting Consistency Regularization for Semi-Supervised Learning}
% You can use \thanks for acknowledgment. Do not add any acknowledgment to the draft 
% version that is used for the review process.  
%\title{Title\thanks{XXX}}

\ifreview
	% ANONYMOUS SUBMISSION FOR REVIEW
	% DO NOT MODIFY these for the draft version that is used for the review process.
	\titlerunning{DAGM GCPR 2021 Submission \SubNumber{}. CONFIDENTIAL REVIEW COPY.}
	\authorrunning{DAGM GCPR 2021 Submission \SubNumber{}. CONFIDENTIAL REVIEW COPY.}
	\author{DAGM GCPR 2021 - \GCPRTrack{}}
	\institute{Paper ID \SubNumber}
\else
	\authorrunning{Yue Fan, Anna Kukleva and Bernt Schiele}
    \author{Yue Fan \qquad Anna Kukleva \qquad Bernt Schiele\\
    {\tt\small \{yfan, akukleva, schiele\}@mpi-inf.mpg.de 
    }
    }
    \institute{Max Planck Institute for Informatics, Saarbrücken, Germany
    \\ Saarland Informatics Campus
    }

% 	\author{First Author\inst{1}\orcidID{0000-1111-2222-3333} \and
% 	Second Author\inst{2,3}\orcidID{1111-2222-3333-4444} \and
% 	Third Author\inst{3}\orcidID{2222--3333-4444-5555}}
	
% 	\authorrunning{F. Author et al.}
% 	% First names are abbreviated in the running head.
% 	% If there are more than two authors, 'et al.' is used.
	
% 	\institute{Princeton University, Princeton NJ 08544, USA \and Springer Heidelberg, Tiergartenstr. 17, 69121 Heidelberg, Germany
% 	\email{lncs@springer.com}\\
% 	\url{http://www.springer.com/gp/computer-science/lncs} \and ABC Institute, Rupert-Karls-University Heidelberg, Heidelberg, Germany\\
% 	\email{\{abc,lncs\}@uni-heidelberg.de}}
\fi

\maketitle              % typeset the header of the contribution

\begin{abstract}
Consistency regularization is one of the most widely-used techniques for semi-supervised learning (SSL).
Generally, the aim is to train a model that is invariant to various data augmentations.
In this paper, we revisit this idea and find 
% \anna{i'm not sure about the phrasing here, I think that  ``find'' is not the best word. What about ``confirm''? or sth along these lines..}\bernt{`find' works well for me}
that enforcing invariance by decreasing distances between features from differently augmented images leads to improved performance. However, encouraging equivariance instead, by increasing the feature distance, further improves performance.
To this end, we propose an improved consistency regularization framework by a simple yet effective technique, FeatDistLoss, that imposes consistency and equivariance on the classifier and the feature level, respectively. 
Experimental results show that our model defines a new state of the art for various datasets and settings and outperforms previous work by a significant margin, particularly in low data regimes. 
Extensive experiments are conducted to analyze the method, and the code will be published. 

% \keywords{Semi-supervised learning \and Consistency regularization.}
\end{abstract}
\section{Introduction} \label{sec:intro}
Deep learning requires large-scale and annotated datasets to reach state-of-the-art performance \cite{russakovsky2015imagenetchallenge,lin2014coco}. 
As labels are not always available or expensive to acquire %annotations, obtaining unlabeled data is often much simpler. %\anna{unlabeled data is often considered as an additional source of information. }\bernt{I like the previous formulation better as it is the direct contrast to the previous point} 
%Thus, 
a wide range of semi-supervised learning (SSL) methods have been proposed to leverage unlabeled data~\cite{tarvainen2017meanteachers,laine2016pimodel,miyato2018vat,verma2019ict,berthelot2019mixmatch,sohn2020fixmatch,xie2019uda,berthelot2019remixmatch,arazo2020plcb,lee2013pseudo,pham2020meta,french2020milking,bachman2019learning,chen2020simclrv2}.

% Deep learning requires large-scale and annotated datasets to reach state-of-the-art performance \cite{russakovsky2015imagenetchallenge,lin2014coco}. 
% While labels are not always available and it is expensive to acquire annotations for a large-scale dataset, obtaining unlabeled data is often much simpler. %\anna{unlabeled data is often considered as an additional source of information. }\bernt{I like the previous formulation better as it is the direct contrast to the previous point} 
% Many semi-supervised learning (SSL) methods propose to leverage unlabeled data~\cite{tarvainen2017meanteachers,laine2016pimodel,miyato2018vat,verma2019ict,berthelot2019mixmatch,sohn2020fixmatch,xie2019uda,berthelot2019remixmatch,arazo2020plcb,lee2013pseudo,pham2020meta,french2020milking,bachman2019learning,chen2020simclrv2}.

\begin{figure}
    \centering
    \captionsetup{font=footnotesize,labelfont=footnotesize}
    \includegraphics[width=0.48\linewidth]{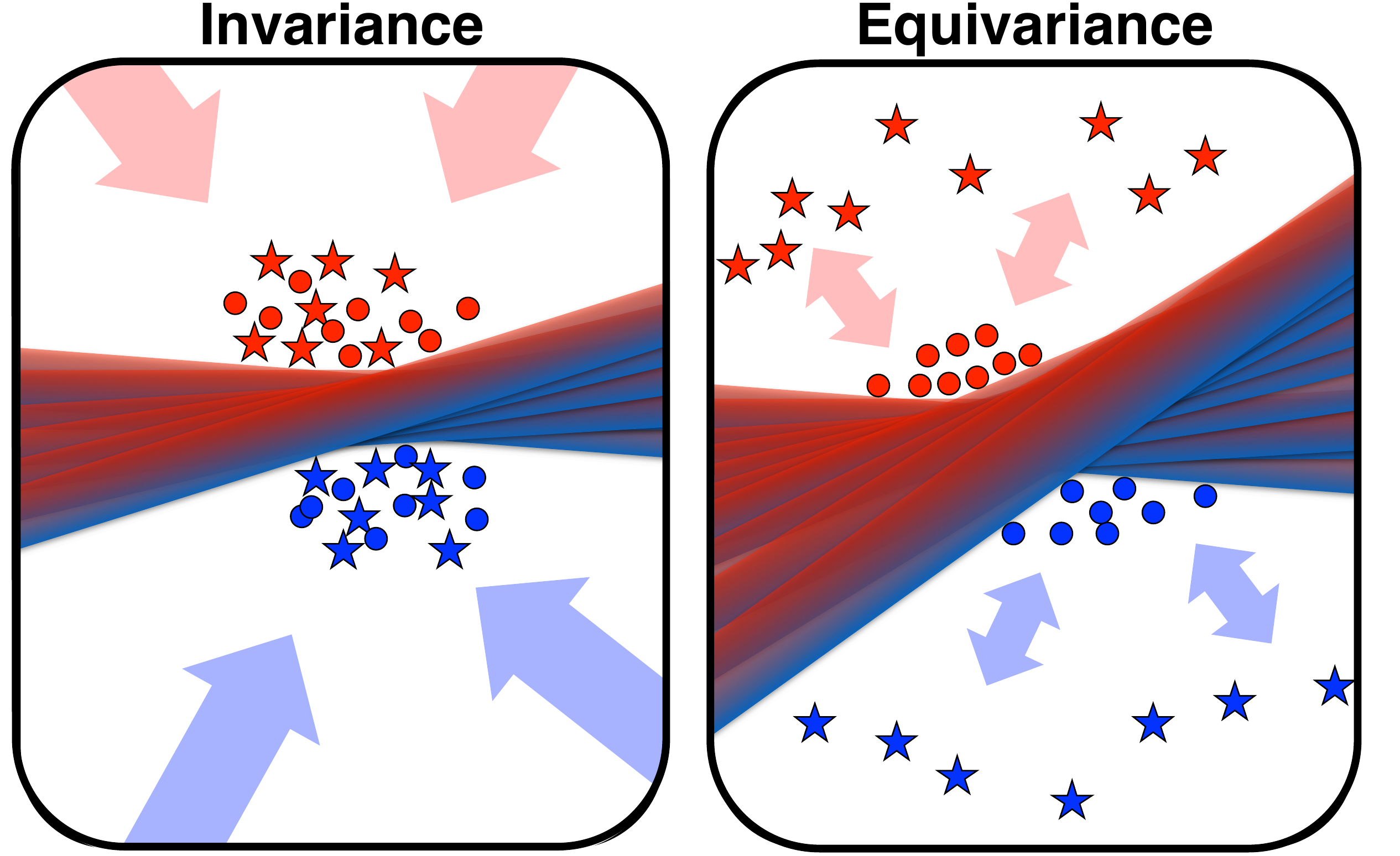}
    \includegraphics[width=0.5\linewidth]{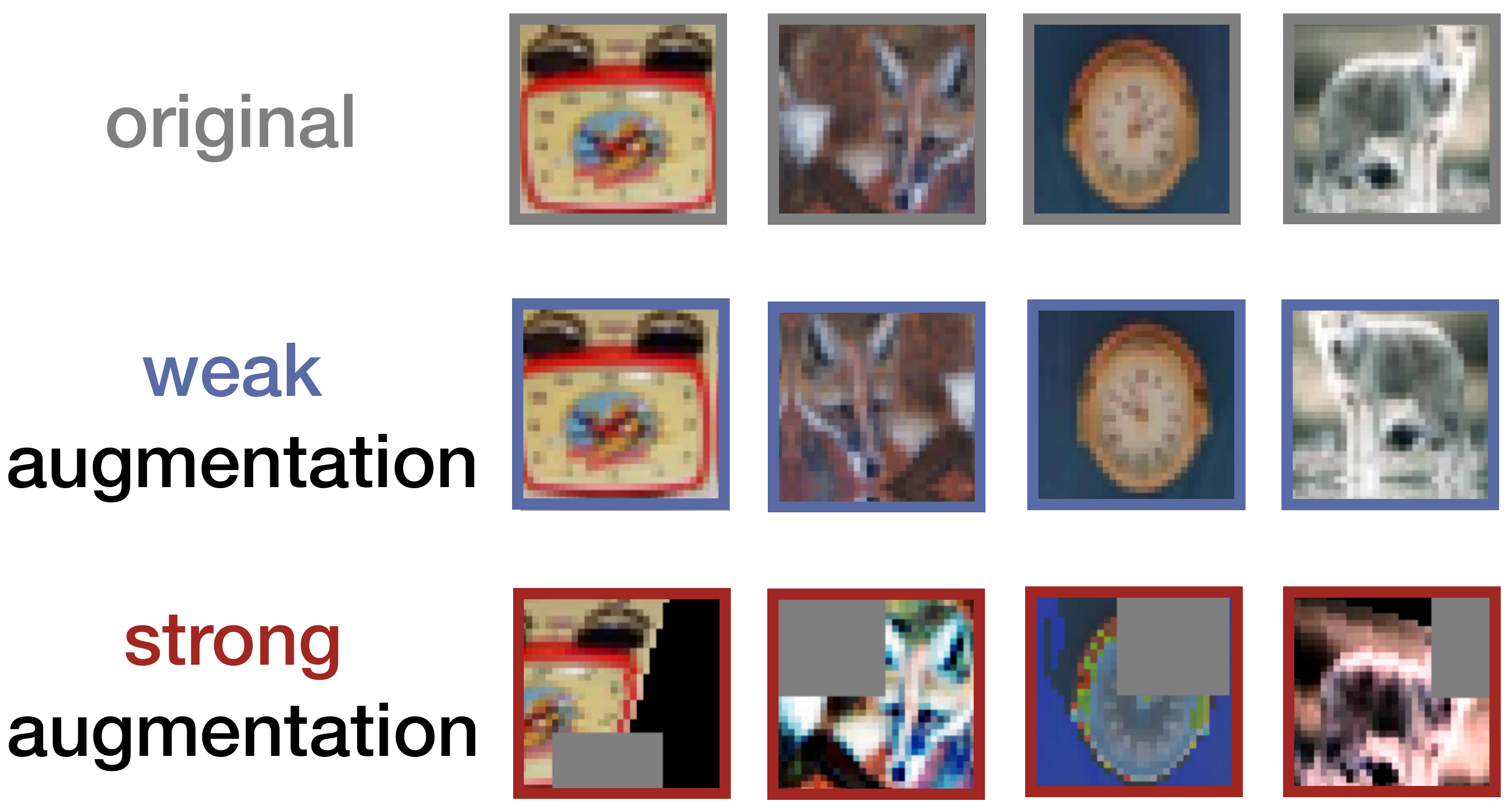}
    \caption{
    \textbf{Left:}
    Binary classification task.
    Stars are features of strongly augmented images and circles are of weakly augmented images.
    %that are augmented with different strengths.
    While encouraging invariance by decreasing distance between features from differently augmented images gives good performance (left), encouraging equivariant representations by increasing the distance regularizes the feature space more, leading to even better generalization performance.
    \textbf{Right:}
    Examples of strongly and weakly augmented images from CIFAR-100. 
    % Original images are taken from CIFAR-100.
    The visually large difference between them indicates that it can be more beneficial if they are treated differently.
    }
    \vspace{-20pt}
    \label{fig:teaser}
\end{figure}

Consistency regularization \cite{bachman2014firstconsistregular,laine2016pimodel,sajjadi2016firstconsistregular} is one of the most widely-used SSL methods. Recent work~\cite{sohn2020fixmatch,xie2019uda,kuo2020featmatch} achieves strong performance by utilizing unlabeled data in a way that model predictions should be invariant to input perturbations. %\bernt{the previous sentence is essentially identical to the one in the abstract that Fan criticized} \fan{since we use the previous formulation, I guess it should also be fine here :)}
% As one of the most widely-used SSL methods, the idea of consistency regularization \cite{bachman2014firstconsistregular, laine2016pimodel, sajjadi2016firstconsistregular} achieved great success~\cite{sohn2020fixmatch, xie2019uda, kuo2020featmatch} by utilizing unlabeled data in a way that model predictions should be invariant to input perturbations.
However, when using advanced and strong data augmentation schemes,  
we question if the model should be invariant to such strong perturbations.
%  \fan{I'm fine with both - @Bernt, which sounds better to you?} it is not obvious that the model should be invariant to such strong perturbations. 
On the right of Figure \ref{fig:teaser} we illustrate that strong data augmentation leads to perceptually highly diverse images. 
% As an illustration see figure \ref{fig:teaser} right, where we show that strong data augmentation can lead to perceptually highly diverse images.
%\anna{
Thus, we argue that improving equivariance on such strongly augmented images can provide even better performance rather than making the model invariant to all kinds of augmentations.
%}
%Thus, we argue that it might make sense to treat such strongly augmented images differently rather than to make the model invariant to all kinds of augmentations.
To this end, we propose a simple yet effective technique, Feature Distance Loss (FeatDistLoss), to improve data-augmentation-based consistency regularization.

%  in combination with existing strong techniques for semi-supervised learning

%%%%%%%%%%%%%%%%%%%%% rewritten partly by Anna (start)

We formulate our FeatDistLoss as to explicitly encourage invariance or equivariance between features from different augmentations while enforcing the same semantic class label. 
Figure \ref{fig:teaser} left shows the intuition behind the idea. %\bernt{figure not yet updated - correct?}
Specifically, encouragement of equivariance for the same image but different augmentations (increase distance between stars and circles of the same color) pushes representations apart from each other, thus, covering more space for the class. 
Imposing invariance, on the contrary, makes the representations of the same semantic class more compact. 
In this work we empirically find that increasing equivariance to differently augmented versions of the same image can lead to better performance especially when rather few labels are available per class (see section \ref{sec:analysis}). 

This paper introduces the method \textit{CR-Match} which combines FeatDistLoss with other strong techniques defining a new state-of-the-art across a wide range of settings of standard SSL benchmarks, including CIFAR-10, CIFAR-100, SVHN, STL-10, and Mini-Imagenet.
%
%Our code will be made publicly available.
%
More specifically, our contribution is fourfold.
%\begin{itemize}
%  \item 
(1) We improve data-augmentation-based consistency regularization by a simple yet effective technique for SSL called \textit{FeatDistLoss} which regularizes the distance between feature representations from differently augmented images of the same class.
%  \item 
(2) We show that while encouraging invariance results in good performance, encouraging equivariance to differently augmented versions of the same image  consistently results in even better generalization performance.
%   \item While encouraging invariance by decreasing the feature distance during learning results in good performance, encouraging equivariance by increasing the feature distance consistently results in better generalization performance.
%  \item 
(3) We provide comprehensive ablation studies on different distance functions and different augmentations with respect to the proposed FeatDistLoss. 
%   \item We provide comprehensive ablation studies to understand the proposed FeatDistLoss by exploring different ways of applying it, including different distance functions, different augmentations, and increasing or decreasing the feature distance.
%  \item 
(4) In combination with other strong techniques, we achieve \textit{new state-of-the-art results} across a variety of standard semi-supervised learning benchmarks, specifically in low data regimes. 

\section{Related Work}
\vspace{-5pt}
% \bernt{related work looks very short - we mihgt want to consider to expand once the rest of the paper is in better shape}
% \bernt{since SSL is part of our story we should actually start with a brief overview of SSL-methods. While we cannot review all SSL methods obviously, we should briefly talk about those that are relevant to us and in particular those that we compare to in the experiments - and also how we are different. In my opinion we should start with that. At the moment there is too much emphasis on regularization - this can be shortened imho. This will also increase the difference to prior work as we are doing more regularization than most other SSL methods I'd say - tbd}
SSL is a broad field %with a huge diversity of methods, 
aiming to exploit both labeled and unlabeled data.
%exploit unlabeled data to improve performance on labeled data.
Consistency regularization is a powerful method for SSL~\cite{rasmus2015ladder,sajjadi2016firstconsistregular,bachman2014firstconsistregular}.
The idea is that the model should output consistent predictions for perturbed versions of the same input.
% \anna{Many works explore different ways to generate such perturbations that help to preserve essential information for various down-stream tasks. }
Many works explored different ways to generate such perturbations.
%that help to preserve essential information.
% Methods have been proposed to generate perturbations in different ways.
For example, \cite{tarvainen2017meanteachers} uses an exponential moving average of the trained model to produce another input; \cite{sajjadi2016firstconsistregular,laine2016pimodel} use random max-pooling and Dropout~\cite{srivastava2014dropout}; \cite{xie2019uda,berthelot2019remixmatch,sohn2020fixmatch,kuo2020featmatch} use advanced data augmentation; \cite{berthelot2019mixmatch,verma2019ict,berthelot2019remixmatch} use MixUp regularization \cite{zhang2017mixup}, which encourages convex behavior ``between'' examples.
Another spectrum of popular approaches is pseudo-labeling~\cite{scudder1965firstpl,nesterov27firstpl,lee2013pseudo}, % or self-training \cite{rosenberg2005selftrain}, 
where the model is trained with artificial labels. 
% \anna{the next sentence is completely boring, gives nothing. Better remove words entropy minimization - it's jsut usual classification, how you do that - does not matter here, better talk about how these pseudo-labels can be generated}
\cite{arazo2020plcb} trained the model with ``soft'' pseudo-labels from network predictions; \cite{pham2020meta} proposed a meta learning method that deploys a teacher model to adjust the pseudo-label alongside the training of the student;
\cite{sohn2020fixmatch,lee2013pseudo} learn from ``hard'' 
% \bernt{please use ``bla'' rather than "bla" - one of these latex things - please check everywhere in the text} 
pseudo-labels and only retain a pseudo-label if the largest class probability is above a predefined threshold.
% As an implicit form of entropy minimization \cite{grandvalet2005minentropy}, pseudo-labeling has been employed as a part of many successful methods and achieved good results \cite{lee2013pseudo,rosenberg2005selftrain,arazo2020plcb,sohn2020fixmatch,pham2020meta}.
Furthermore, there are many excellent works around generative models \cite{kingma2014deepgan,odena2016gan,denton2016cgan} and graph-based methods \cite{luo2018smoothgraph,liu2019deepgraph,bengio200611graph,joachims2003transductivegraph}. 
%, which are less related to our work, and thus not discussed in detail here.
We refer to \cite{chapelle2009semi,zhu05semisurvey,zhu2009semiintro} for a more comprehensive introduction of SSL methods.
% A more comprehensive introduction of SSL methods is available in~\cite{chapelle2009semi,zhu05semisurvey,zhu2009semiintro}.

% where they added an additional loss that minimizes the entropy of the model’s predicted class distribution.
% An implicit form of entropy minimization is pseudo-labeling \cite{lee2013pseudo} or self-training \cite{rosenberg2005selftrain}, which has been widely employed in recent methods and gives good performance \cite{arazo2020plcb, sohn2020fixmatch, kuo2020featmatch}.

% Next, we discuss methods that are most relevant to CR-Match and make a contrast to our work. 
% As is underlined by \cite{xie2019uda}, noise injection plays a crucial role in consistency regularization, and  advanced data augmentation methods, especially a combination of weak and strong augmentations, to noise unlabeled data brings substantial improvements \cite{berthelot2019remixmatch, sohn2020fixmatch}.
Noise injection plays a crucial role in consistency regularization~\cite{xie2019uda}. Thus advanced data augmentation, especially  combined with weak data augmentation, introduces stronger noise to unlabeled data and brings substantial improvements \cite{berthelot2019remixmatch,sohn2020fixmatch}.
% \bernt{the previous sentence was off English-wise - can you please check if the previous sentences make sense as they are now?}
\cite{sohn2020fixmatch} proposes to integrate pseudo-labeling into the pipeline by computing pseudo-labels from weakly augmented images, and then uses the cross-entropy loss between the pseudo-labels and strongly augmented images.
Besides the classifier level consistency, our model also introduces consistency on the feature level, which explicitly regularizes representation learning and shows improved generalization performance.
% and then optimizing with respect to the corresponding strongly augmented images.
% While using the consistency regularization technique, CR-Match further deploys a novel feature distance loss to directly regularize the feature space, which is shown by extensive experiments to be superior over other methods in terms of performance. 
% \anna{While using the consistency regularization technique, CR-Match explores properties of the feature space by means of novel regularization technique applying feature distance loss to directly influence decision boundaries . We also shown by extensive experiments to be superior over other methods in terms of performance.} 
Moreover, self-supervised learning is known to be beneficial in the context of SSL.
In \cite{he2020moco,chen2020simclr,chen2020simclrv2,rebuffi2020semiinlow}, self-supervised pre-training is used to initialize SSL.
However, these methods normally have several training phases, where many hyper-parameters are involved.
We follow the trend of \cite{zhai2019s4l,berthelot2019remixmatch} to incorporate an auxiliary self-supervised loss alongside training. 
Specifically, we optimizes a rotation prediction loss \cite{gidaris2018rotnet}.

Equivariant representations are recently explored by capsule networks \cite{capsule2018sabour,capsule2018hinton}.
They replaced max-pooling layers with convolutional strides and dynamic routing to preserve more information about the input, allowing for preservation of part-whole relationships in the data. 
It has been shown, that the input can be reconstructed from the output capsule vectors.
Another stream of work on group equivariant networks~\cite{cohen2016group, NEURIPS2019_45d6637b, cohen2016steerable} explores various equivariant architectures that produce transform in a predictable linear manner under transformations of the input.
Different from previous work, our work explores equivariant representations in the sense that differently augmented versions of the same image are represented by different points in the feature space despite the same semantic label.
% Our work explores equivariant representations in a different way, where we increase the feature distance between images from different augmentations while imposing the same class label.
As we will show in section \ref{sec:analysis}, information like  object location or orientation is more predictable from our model when features are pushed apart from each other.

\section{CR-Match} \label{sec:method}
\vspace{-5pt}
% \bernt{the term CR-Match is not introduced in that intro - should be done obviously}
% \bernt{currently this section is not particularly exciting - very dry and descriptive rather than giving intuitions why the different terms are making sense and what alternatives we can/should/have considered.}
% As a highly-successful and widely-adopted methodology in SSL \cite{bachman2014firstconsistregular, laine2016pimodel, sajjadi2016firstconsistregular, sohn2020fixmatch, xie2019uda, kuo2020featmatch}, we aim to leverage and further improve the consistency regularization framework.
Consistency regularization is highly-successful and widely-adopted technique in SSL \cite{bachman2014firstconsistregular,laine2016pimodel,sajjadi2016firstconsistregular,sohn2020fixmatch,xie2019uda,kuo2020featmatch}. 
% In this work, we aim to leverage and improve it by further increasing regularization of the feature space.  
In this work, we aim to leverage and improve it by even further regularizing the feature space.  
To this end, we present a simple yet effective technique FeatDistLoss to explicitly regularize representation learning and classifier learning at the same time. 
We describe our SSL method, called CR-Match, which shows improved performance across many different settings, especially in scenarios with few labels.  %\anna{I'd remove the above sentence and then ``In this section ..''}
In this section, we first describe our technique FeatDistLoss and then present CR-Match that combines FeatDistLoss with other regularization techniques inspired from the literature. %\anna{with other strong techniques inspired from the literature.} \fan{rotnet is actually a regularization. We were trying to sell the regularization story back then :)}

% \begin{figure}[!]
%     \centering
%     % \includegraphics[width=1\textwidth]{weak_strong_DA_row.pdf}
%     % \includegraphics[width=\linewidth]{weak_strong_DA_col.pdf}
%     \includegraphics[width=0.9\linewidth]{augm.pdf}
%     \caption{Examples of strongly and weakly augmented images. Original images are taken from CIFAR-100. 
%     % Weak augmentation refers to random cropping and random mirroring with probability 0.5. Strong augmentation is a combination of RandAugment \cite{cubuk2020randaugment} and CutOut \cite{devries2017cutout}. \fan{not sure if we should write out details of weak and strong augmentations}
%     The visually large difference between weakly and strongly augmented images indicates that it can be more beneficial if they are treated differently.
%     % \bernt{figure was FAR too big - the data augmentations are not a contribution of this paper - I rescaled the figure and that would still work perfectly in terms of image size for the purpose of illustration - I would not go larger - rather smaller}
%     }
%     \label{fig:weak_strong_DA}
% \end{figure}
\vspace{-5pt}
\subsection{Feature Distance Loss} 

\begin{figure}[!]
    \centering
    \captionsetup{font=footnotesize,labelfont=footnotesize}
    \vspace{-10pt}
    \includegraphics[width=\linewidth]{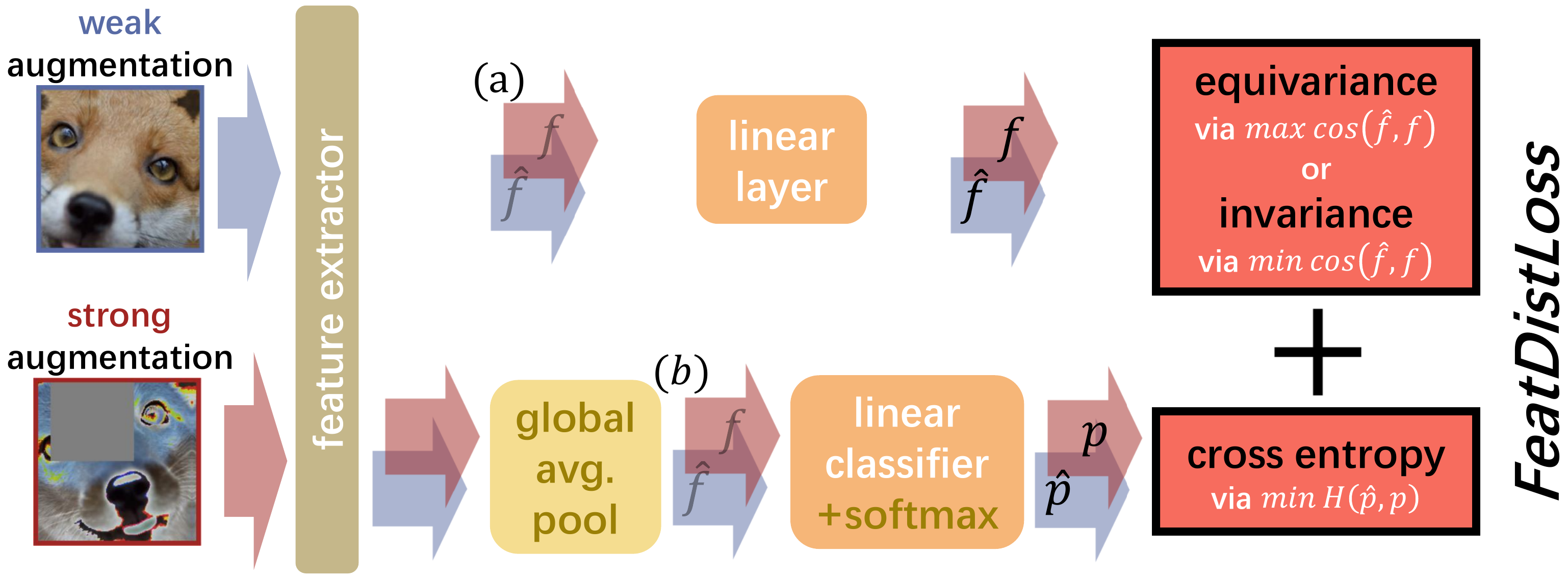}
    \caption{
    %\bernt{in the figure I would replace `entropy' with `cross entropy'}
    The proposed FeatDistLoss utilizes unlabeled images in two ways: %\anna{The proposed FeatDistLoss utilizes unlabeled images in two ways: on the classifier level, labeled and unlabeled images should belong to one of the predefined classes, whereas on the feature level we enforce additional regularization to help the classifier through increasing equivariance or invariance. } \fan{it is not ``labeled and unlabeled image'' but ``versions of the same image'' that belong to the same class.}
    % The proposed architecture performs losses on two levels. 
    On the classifier level, different versions of the same image should generate the same class label, whereas on the feature level, representations are encouraged to become either more equivariant (pushing away) or invariant (pulling together). 
    $f$ and $\hat{f}$ denote strong and weak features; $p$ and $\hat{p}$ are predicted class distributions from strong and weak features; a) and b) denote features before and after the global average pooling layer.
    Our final model takes features from a) and encourages equivariance to differently augmented versions of the same image. An ablation study of other choices is in section \ref{sec:analysis}.
    % \bernt{The figure is not ideal in the following sense: FeatDistLoss is used as the term to refer to both types of losses - but then in the right branch we should not have `FeatDistLoss' written but something else that cannot be confused with it - canonical would be `feature distance'} 
    % \bernt{another comment: the figure is visually dominated by the orange and yellow boxes - even though the red boxes are more important - please change the figure such that the red boxes are indeed the salient part of the figure}
    }
    \vspace{-20pt}
    \label{fig:arch}
\end{figure}

\myparagraph{Background:}
% \anna{Methods that employ consistency regularization \cite{bachman2014firstconsistregular,laine2016pimodel,sajjadi2016firstconsistregular} encourage the model parameters to make the predictions of the model invariant to input perturbations.  
% }
The idea of consistency regularization \cite{bachman2014firstconsistregular,laine2016pimodel,sajjadi2016firstconsistregular} is to encourage the model predictions to be invariant to input perturbations.
Given a batch of $n$ unlabeled images $\textbf{u}_i, i \in (1, ..., n)$, consistency regularization can be formulated as the following loss function:
\begin{equation}
\frac{1}{n} \sum_{i=1}^{n} ||f(\mathcal{A}(\textbf{u}_i)) - f(\alpha(\textbf{u}_i)) ||^2_2 \label{lcr}
\end{equation}
where $f$ is an encoder network that maps an input image to a $d$-dimensional feature space; $\mathcal{A}$ and $\alpha$ are two stochastic functions which are, in our case, strong and weak augmentations, respectively (details in Section \ref{sec:detail}). 
By minimizing the $L_2$ distance between perturbed images, the representation is therefore encouraged to become more invariant with respect to different augmentations, which helps  generalization.
The intuition behind this is that a good model should be robust to data augmentations of the images.
% \bernt{the above essentially just gives the equation without any real intuition what that might sense - please add a few sentences about the intuition and in particular mention that this encourages the representation to become more invariant w.r.t. the augmentations}

\myparagraph{FeatDistLoss:}
As shown in Figure \ref{fig:arch}, %\bernt{what I a missing is a discussion about the different options how to perform FeatDistLoss as shown in figure 3 - figure 3 seems to be referred to just once - I would strongly encourage to add a few sentences about the potential choices probably after equation 2 to briefly discuss options and to mention that we are performing experiments about this and that (a) was found to work best experimentally - if we do not have a discussion about this in this section it might be lost for many readers as they can miss this in the experiments too easily} \fan{I agree. I added a line after equation 2, please check.} 
we extend the above consistency regularization idea by introducing consistency on the classifier level and invariance or equivariance on the feature level. %introducing consistency regularization on both the classifier and the feature level.
FeatDistLoss thus allows to apply different types of control for these levels. In particular, when encouraging to reduce the feature distance, it becomes similar to classic consistency regularization, and encourages invariance between differently augmented images.
As argued above, making the model predictions invariant to input perturbations gives good generalization performance. Instead, in this work we find it is more beneficial to treat images from different augmentations differently because some distorted images are largely different from their original images as demonstrated visually in Figure \ref{fig:teaser} right. %\fan{we need more equivariance story here. I think we need to let the reader understand ``we will use CR-Equiv for the main results and CR-Inv will only be used for ablation study".}\bernt{I don't think we need more story here - there is quite some text already - here and in the previous sections - you might want to introduce the notation here for CR-Equiv and CR-Inv -- btw: I would have used the abbreviations with the `.' at the end -- none of other abbreviations has a `.' at the end in this paper} 
Therefore, the final model (CR-Match) uses FeatDistLoss to increase the distance between image features from augmentations of different intensities while at the same time enforcing the same semantic label for them.
Note that in Section~\ref{sec:analysis}, we conduct an ablation study on the choice of distance function, where we denote CR-Match as CR-Equiv, and the model that encourages invariance as CR-Inv.
% In section \ref{sec:ablation}, we denote models that encourage invariance and equivariance by CR-Inv and CR-Equiv, respectively. 
% \fan{I think readers will be confused about the difference between CR-Match and CR-Equiv (CR-Match = CR-Equiv - RotNet). But actually I don't want to draw their attention to this subtle difference, to avoid potential questions about the role of RotNet in our model. Any suggestions?} \bernt{I agree that this is not ideal. I am not concerned about RotNet - but the other part. At some point I was thinking about calling this CR-Match-Inc and CR-Match-Equ and then mentioning that CR-Match is just CR-Match-Equ - but this is probably not much better as it still uses two terms for the same thing...}  \fan{What about ``Note that, in section 4.2, we conduct an ablation study on the choice of distance function, where we denote CR-Match as CR-Equiv, and the model that encourages invariance as CR-Inv''?}\bernt{slightly modified - works for me now}
%\bernt{yet again `only' pushing away and not both pushing away and pulling together implicitly thought of in the previous sentence}

%Importantly, FeatDistLoss obtains better performance by increasing the feature distance between images from augmentations of different intensities while imposing the same class label. %\bernt{yet again `only' pushing away and not both pushing away and pulling together implicitly thought of in the previous sentence}

The final objective for the FeatDistLoss consists of two terms: $\mathcal{L}_{Dist}$ (on the feature level), that explicitly regularizes feature distances between embeddings, and a standard cross-entropy loss $\mathcal{L}_{PseudoLabel}$ (on the classifier level) based on pseudo-labeling.

With $\mathcal{L}_{Dist}$ we either decrease or increase the feature distance 
%\bernt{yet again `only' pushing away and not both pushing away and pulling together implicitly thought of in the previous sentence} 
between weakly and strongly augmented versions of the same image in a low-dimensional space projected from the original feature space to overcome the curse of dimensionality \cite{bellman1966curse}. 
Let $d(\cdot, \cdot)$ be a distance metric and $z$ be a linear layer that maps the high-dimensional feature into a low-dimensional space.
Given an unlabeled image $\textbf{u}_i$, we first extract features with strong and weak augmentations by $f(\mathcal{A}(\textbf{u}_i))$ and $f(\alpha(\textbf{u}_i))$ as  shown in Figure~\ref{fig:arch} (a), and then  FeatDistLoss is computed as:
\begin{equation}
\mathcal{L}_{Dist}(\textbf{u}_i) = d(z(f(\mathcal{A}(\textbf{u}_i))),z(f(\alpha(\textbf{u}_i))) ) \label{ld}
\end{equation}
Different choices of performing $\mathcal{L}_{Dist}$ are studied in Section \ref{sec:analysis}, where we find empirically that applying $\mathcal{L}_{Dist}$ at (a) in Figure \ref{fig:arch} gives the best performance.
% \bernt{I do not understand the part of `features extracted at figure~\ref{fig:arch} (b) into (a)' - I'd rather just say applying the FeatDistLoss at (a) in Figure 3 or something like that...}

At the same time, images from strong and weak augmentations should have the same class label because they are essentially generated from the same original image.
Inspired by \cite{sohn2020fixmatch}, given an unlabeled image $\textbf{u}_i$, a pseudo-label distribution is first generated from the weakly augmented image by $\hat{\textbf{p}}_i=g(f(\alpha(\textbf{u}_i)))$, and then a cross-entropy loss is computed between the pseudo-label and the prediction for the corresponding strongly augmented version as:
\begin{equation}
\mathcal{L}_{PseudoLabel}(\textbf{u}_i) = \ell_{CE}(\hat{\textbf{p}}_i, g(f(\mathcal{A}(\textbf{u}_i)))) \label{lpl}
\end{equation}
where $\ell_{CE}$ is the cross-entropy, $g$ is a linear classifier that maps a feature representation to a class distribution, and $\mathcal{A}(\textbf{u}_i)$ denotes the operator for strong augmentations. % \anna{$\mathcal{A}(\cdot)$ denotes operator for strong augmentation.}

Putting it all together, FeatDistLoss
% \bernt{this is not yet CR-Match - right? Shouldn't we argue taht this is FestDistLoss? at least that would be consistent with fig 2} 
processes a batch of unlabeled data $\textbf{u}_i, i \in (1, ..., B_u)$ with the following loss:
\begin{equation}
\mathcal{L}_{U} = \frac{1}{B_u} \sum_{i=1}^{B_u} \mathbbm{1}\{c_i>\tau\} (\mathcal{L}_{Dist}(\textbf{u}_i) + \mathcal{L}_{PseudoLabel}(\textbf{u}_i))\label{lu}
\end{equation}
where $c_i=max \ \hat{\textbf{p}}_i$ is the confidence score, and $\mathbbm{1}\{\cdot\}$ is the indicator function which outputs 1 when the confidence score is above a threshold.
This confidence thresholding mechanism ensures that the loss is only computed for unlabeled images for which the model generates a high-confidence prediction, which gives a natural curriculum to balance labeled and unlabeled losses \cite{sohn2020fixmatch}.

As mentioned before, depending on the function $d$, FeatDistLoss can decrease the distance between features from different data augmentation schemes (when $d$ is a distance function, thus pulling the representations together), or increase it (when $d$ is a similarity function, thus pushing the representations apart). 
As shown in table \ref{tab:analysis_featdistloss}, we find that both cases results in an improved performance. 
However, increasing the distance between weakly and strongly augmented examples consistently results in better generalization performance. % \anna{shows better generalization.}
We conjecture that the reason lies in the fact that FeatDistLoss by increasing the feature distance explores equivariance properties (differently augmented versions of the same image having distinct features but the same label) of the representations. It encourages the model to have more distinct weakly and strongly augmented images while still imposing the same label, which leads to both more expressive representation and more powerful classifier.
As we will show in Section \ref{sec:analysis}, information like object 
location or orientation is more predictable from models trained with FeatDistLoss that pushes the representations apart.
Additional ablation studies of other design choices such as the distance function and the linear projection $z$ are also provided in Section \ref{sec:analysis}.

\vspace{-5pt}
\subsection{Overall CR-Match}
% \bernt{can we call this section `Overall CR-March Method' or `Complete CR-Match Model' or something like that - from the title it is unclear that the complete model is described in this section}
% \bernt{I personally would prefer if the first paragraph of this section were at the beginning of section 3 - cross-entropy and the rotation prediction loss should stay here - but starting with an adapted paragraph that introduces all three losses and then focuses in section 3.1  on $\mathcal{L}_{U}$ would be more convincing to me.} 
Now we describe our SSL method called CR-Match leveraging the above FeatDistLoss.
Pseudo-code for processing a batch of labeled and unlabeled examples is
shown in algorithm 1 in supplementary material.

Given a batch of labeled images with their labels as $\mathcal{X}=\{(\textbf{x}_i, \textbf{p}_i):i\in(1, ..., B_s)\}$ and a batch of unlabeled images as $\mathcal{U}=\{\textbf{u}_i:i\in(1, ..., B_u)\}$.
CR-Match minimizes the following learning objective:
\begin{equation}
% \resizebox{0.48\textwidth}{!}{
\mathcal{L}_{S}(\mathcal{X}) + \lambda_{u} \mathcal{L}_{U}(\mathcal{U}) + \lambda_{r} \mathcal{L}_{Rot}(\mathcal{X} \cup \mathcal{U}) \label{l}
% }
\end{equation}
where $\mathcal{L}_{S}$ is the supervised cross-entropy loss for labeled images with weak data augmentation regularization;
$\mathcal{L}_{U}$ is our novel feature distance loss for unlabeled images which explicitly regularizes the distance between weakly and strongly augmented images in the feature space;
and $\mathcal{L}_{Rot}$ is a self-supervised loss for unlabeled images and stands for rotation prediction from \cite{gidaris2018rotnet} to provide an additional supervisory and regularizing signal.

\myparagraph{Fully supervised loss for labeled data:} 
We use cross-entropy loss with weak data augmentation regularization for labeled data: 
\begin{equation}
\mathcal{L}_{S} = \frac{1}{B_s} \sum_{i=1}^{B_s} \ell_{CE}(\textbf{p}_i, g(f(\alpha(\textbf{x}_i)))) \label{lx}
\end{equation}
where $\ell_{CE}$ is the cross-entropy loss, $\alpha(\textbf{x}_i)$ is the extracted feature from a weakly augmented image $\textbf{x}_i$, $g$ is the same linear classifier as in equation \ref{ld}, and $\textbf{p}_i$ is the corresponding label for $\textbf{x}_i$.

\myparagraph{Self-supervised loss for unlabeled data:}
Rotation prediction~\cite{gidaris2018rotnet} (RotNet) is one of the most successful self-supervised learning methods, and has been shown to be complementary to SSL methods \cite{zhai2019s4l,berthelot2019mixmatch,rebuffi2020semiinlow}. % \anna{give more citations}
% We integrate the rotation prediction loss in our training framework with a predictor head $h$ that takes as input the representation from the encoder $f$ and outputs the rotation. 
Here, we create four rotated images by $0^{\circ}$, $90^{\circ}$, $180^{\circ}$, and $270^{\circ}$ for each unlabeled image $\textbf{u}_i$ for $i\in(1, ..., \mu B)$.
Then, classification loss is applied to train the model predicting the rotation as a four-class classification task:
\begin{equation}
\mathcal{L}_{Rot} = \frac{1}{4 B_u} \sum_{i=1}^{B_u} \sum_{r \in \mathbbm{R}} \ell_{CE}(r, h(\alpha(Rotate(\textbf{u}_i,r)))) \label{lr}
\end{equation}
where $\mathbbm{R}$ is $\{0^{\circ},90^{\circ},180^{\circ},270^{\circ}\}$, and $h$ is a predictor head.

\vspace{-5pt}
\subsection{Implementation Details} \label{sec:detail}
\myparagraph{Data augmentation:} \label{p:da}
As mentioned above, CR-Match adopts two types of data augmentations: weak augmentation and strong augmentation from \cite{sohn2020fixmatch}.
Specifically, the weak augmentation $\alpha$ corresponds to a standard random cropping and random mirroring with probability $0.5$, and the strong augmentation $\mathcal{A}$ is a combination of RandAugment \cite{cubuk2020randaugment} and CutOut \cite{devries2017cutout}.
At each training step, we uniformly sample two operations for the strong augmentation from a collection of transformations and apply them with a randomly sampled magnitude from a predefined range.
The complete table of transformation operations for the strong augmentation is provided in the supplementary material.
% \bernt{I added the last half-sentence - please check}

% \bernt{mentioning this work is ok- but then it should be clear what exactly is taken from them - right now it reads as if EVERTHING is taken from them - that is neither true nor helpful - you could stat describing all parameters that are NOT taken from fixmatch and then - by cing fixmatch - mention other hyperparameters are traken from them and then describe those - does that make sense?}
% \fan{could you please check if it is ok now?}
\myparagraph{Other implementation details:}
For our results in Section \ref{sec:experiment}, we minimize the cosine similarity in FeatDistLoss, and use a fully-connected layer for the projection layer $z$.
The predictor head $h$ in rotation prediction loss consists of two fully-connected layers and a ReLU \cite{nair2010relu} as non-linearity.
We use the same $\lambda_{u}=\lambda_{r}=1$ in all experiments.
As a common practice, we repeat each experiment with five different data splits and report the mean and the standard deviation of the error rate.
We refer to supplementary material for further details and ablation of the hyper-parameters.
% For our results in section \ref{sec:experiment}, we use the following default hyper-parameters unless otherwise stated.
% We minimize the cosine similarity in FeatDistLoss, and use a fully-connected layer for the projection layer $z$, which maps the feature from the original 8192-dimension space into a 128-dimension space, the same dimension as the feature dimension for classification.
% The predictor head $h$ in rotation prediction loss consists of two fully-connected layers and a ReLU \cite{nair2010relu} as non-linearity.
% We use the same $\lambda_{u}=\lambda_{r}=1$ in all experiments since CR-Match shows good robustness within a range of loss weights in our preliminary experiments.
% We train our model for 512 epochs on CIFAR-10, CIFAR-100, and SVHN.
% On STL-10 and Mini-ImageNet, we train the model for 300 epochs.
% Other hyper-parameters are from \cite{sohn2020fixmatch} for the compatibility.
% Specifically, the confidence thresholds $\tau$ for pseudo-label selection is 0.95.
% We use SGD with momentum 0.9 and cosine learning rate schedule from \cite{sohn2020fixmatch} starting from 0.03, batch size $B_s$ is 64 for labeled data, and $B_u$ is $7 \times B_s$.
% Finally, we report final performance using an exponential moving average of model parameters as recommended by~\cite{tarvainen2017meanteachers}.
% As a common practice, we repeat each experiment with five different data splits and report the mean and the standard deviation of the error rate.

\section{Experimental Results} \label{sec:experiment}
\vspace{-5pt}
% \bernt{it would be could to start with something like: `We follow precious [citations] and conduct...} 
Following protocols from previous work \cite{berthelot2019mixmatch,sohn2020fixmatch}, we conduct experiments on several commonly used SSL
image classification benchmarks to test the efficacy of CR-Match.
We show our main results in Section \ref{sec:main_results}, where we achieve state-of-the-art error rates across all settings on SVHN~\cite{netzer2011svhn}, CIFAR-10~\cite{krizhevsky2009cifar}, CIFAR-100~\cite{krizhevsky2009cifar}, STL-10~\cite{coates2011stl10}, and mini-ImageNet~\cite{ravi2016miniImagenet}. 
In our ablation study in Section \ref{sec:ablation} we analyze the effect of FeatDistLoss and RotNet across different settings.
Finally, in Section \ref{sec:analysis} we extensively analyse various design choices for our FeatDistLoss.

\begin{table*}
\captionsetup{font=scriptsize,labelfont=scriptsize}
\vspace{-10pt}
\resizebox{\textwidth}{!}{%
\begin{tabular}{lrrrrrrrrr}
\toprule
& \multicolumn{3}{c}{CIFAR-10} & \multicolumn{3}{c}{CIFAR-100} \\
        \cmidrule(l{3pt}r{3pt}){1-1} \cmidrule(l{3pt}r{3pt}){2-4}  \cmidrule(l{3pt}r{3pt}){5-7} \cmidrule(l{3pt}r{3pt}){8-10}
        Per class labels & 4 labels & 25 labels & 400 labels & 4 labels & 25 labels & 100 labels \\
        \cmidrule(l{3pt}r{3pt}){1-1} \cmidrule(l{3pt}r{3pt}){2-4}  \cmidrule(l{3pt}r{3pt}){5-7} \cmidrule(l{3pt}r{3pt}){8-10}
Mean Teacher~\cite{tarvainen2017meanteachers} & \multicolumn{1}{c}{-} & 32.32$\pm$2.30$^*$ & 9.19$\pm$0.19$^*$ & \multicolumn{1}{c}{-} & 53.91$\pm$0.57$^*$ & 35.83$\pm$0.24$^*$ \\
MixMatch~\cite{berthelot2019mixmatch} & 47.54$\pm$11.50$^*$ & 11.08$\pm$0.87\ \ & 6.24$\pm$0.06\ \ & 67.61$\pm$1.32$^*$ & 39.94$\pm$0.37$^*$ & 25.88$\pm$0.30\ \ \\
UDA~\cite{xie2019uda} & 29.05$\pm$5.93$^*$ & 5.43$\pm$0.96\ \ & 4.32$\pm$0.08$^*$ & 59.28$\pm$0.88$^*$ & 33.13$\pm$0.22$^*$ & 24.50$\pm$0.25$^*$ \\
ReMixMatch~\cite{berthelot2019remixmatch} & 19.10$\pm$9.64$^*$\ & 6.27$\pm$0.34\ \ & 5.14$\pm$0.04\ \ & \textit{44.28}$\pm$2.06$^*$ & \textit{27.43}$\pm$0.31$^*$ & 23.03$\pm$0.56$^*$ \\
FixMatch (RA)~\cite{sohn2020fixmatch} & 13.81$\pm$3.37\ \ & \textit{5.07}$\pm$0.65\ \ & 4.26$\pm$0.05\ \ & 48.85$\pm$1.75\ \ & 28.29$\pm$0.11\ \ & \textit{22.60}$\pm$0.12\ \ \\
FixMatch (CTA)~\cite{sohn2020fixmatch} & \textit{11.39}$\pm$3.35\ \ & \textit{5.07}$\pm$0.33\ \ & \textit{4.31}$\pm$0.15\ \ &  49.95$\pm$3.01\ \ & 28.64$\pm$0.24\ \ & 23.18$\pm$0.11\ \ \\
FeatMatch~\cite{kuo2020featmatch} & \multicolumn{1}{c}{-} & 6.00$\pm$0.41\ \ & 4.64$\pm$0.11\ \ & \multicolumn{1}{c}{-} & \multicolumn{1}{c}{-} & \multicolumn{1}{c}{-} \\ \hline
CR-Match & \textbf{10.70}$\pm$2.91\ \ & \textbf{5.05}$\pm$0.12\ \ & \textbf{3.96}$\pm$0.16\ \ & \textbf{39.45}$\pm$1.69\ \ & \textbf{25.43}$\pm$0.14\ \ & \textbf{20.40}$\pm$0.08\ \ \\ \bottomrule
\end{tabular}
}%
\vspace{1pt}
\caption{
% \bernt{for cifar-10 you have a 400 labels case - this is wrong? or not?} \fan{It is correct. People benchmark CIFAR-10@400 not CIFAR-10@100.}
Error rates on CIFAR-10, and CIFAR-100. 
A Wide ResNet-28-2 \cite{zagoruyko2016wrn} is used for CIFAR-10 and a Wide ResNet-28-8 with 135 filters per layer \cite{berthelot2019mixmatch} is used for CIFAR-100.
We use the same code base as \cite{sohn2020fixmatch} (i.e., same network architecture and training protocol) to make the results directly comparable. $^*$Numbers are generated by \cite{sohn2020fixmatch}. 
$^\dag$Numbers are produced without CutOut.
The best number is in bold and the second best number is in italic.
}
\vspace{-10pt}
\label{tab:main_cifar}
\end{table*}

\begin{table*}
\captionsetup{font=scriptsize,labelfont=scriptsize}
\vspace{-10pt}
\resizebox{0.6\textwidth}{!}{%
\begin{tabular}{lrrrrrr}
\toprule
        & STL-10     & \multicolumn{3}{c}{SVHN} \\
        \cmidrule(l{3pt}r{3pt}){1-1} \cmidrule(l{3pt}r{3pt}){2-2} \cmidrule(l{3pt}r{3pt}){3-5}
        Per class labels & 100 labels & 4 labels & 25 labels & 100 labels \\
        \cmidrule(l{3pt}r{3pt}){1-1} \cmidrule(l{3pt}r{3pt}){2-2} \cmidrule(l{3pt}r{3pt}){3-5}
Mean Teacher~\cite{tarvainen2017meanteachers} & 21.34$\pm$2.39$^*$ & \multicolumn{1}{c}{-} & 3.57$\pm$0.11$^*$ & 3.42$\pm$0.07$^*$ \\
MixMatch~\cite{berthelot2019mixmatch} & 10.18$\pm$1.46\ \ & 42.55$\pm$14.53$^*$ & 3.78$\pm$0.26\ \ & 3.27$\pm$0.31\ \ \\
UDA~\cite{xie2019uda} & 7.66$\pm$0.56$^*$ & 52.63$\pm$20.51$^*$ & 2.72$\pm$0.40\ \ & \textit{2.23}$\pm$0.07\ \ \\
ReMixMatch~\cite{berthelot2019remixmatch} & 6.18$\pm$1.24\ \  & \textit{3.34}$\pm$0.20$^*$ & 3.10$\pm$0.50\ \ & 2.83$\pm$0.30\ \ \\
FixMatch (RA)~\cite{sohn2020fixmatch} & 7.98$\pm$1.50\ \  & 3.96$\pm$2.17\ \ & \textit{2.48}$\pm$0.38\ \ & 2.28$\pm$0.11\ \ \\
FixMatch (CTA)~\cite{sohn2020fixmatch} & \textit{5.17}$\pm$0.63\ \ & 7.65$\pm$7.65\ \ & 2.64$\pm$0.64\ \ & 2.36$\pm$0.19\ \ \\
FeatMatch~\cite{kuo2020featmatch} & \multicolumn{1}{c}{-} & \multicolumn{1}{c}{-} & 3.34$\pm$0.19$^\dag$ & 3.10$\pm$0.06$^\dag$ \\ \midrule
CR-Match & \textbf{4.89}$\pm$0.17\ \ & \textbf{2.79}$\pm$0.93\ \ & \textbf{2.35}$\pm$0.29\ \ & \textbf{2.08}$\pm$0.07\ \ \\ \bottomrule
\end{tabular}
}%
\resizebox{0.4\textwidth}{!}{%
\begin{tabular}{lrr}
\toprule
& \multicolumn{2}{c}{Mini-ImageNet} \\ 
\cmidrule(l){1-1}  \cmidrule(l){2-3} 
Per class labels & 40 labels & 100 labels \\ 
\cmidrule(l){1-1}  \cmidrule(l){2-3} 
Mean Teacher~\cite{tarvainen2017meanteachers} & 72.51$\pm$0.22 & 57.55$\pm$1.11 \\
Label Propagation~\cite{iscen2019labelpropagation} & 70.29$\pm$0.81 & 57.58$\pm$1.47 \\
PLCB~\cite{arazo2020plcb} & 56.49$\pm$0.51 & 46.08$\pm$0.11 \\
FeatMatch~\cite{kuo2020featmatch} & \textit{39.05}$\pm$0.06 & \textit{34.79}$\pm$0.22 \\ \midrule
CR-Match & \textbf{34.87}$\pm$0.99 & \textbf{32.58}$\pm$1.60 \\ \bottomrule
\end{tabular}
}%
\vspace{1pt}
\caption{
\textbf{Left:}
Error rates on STL-10 and SVHN. 
A Wide ResNet-28-2 and a Wide ResNet-37-2 \cite{zagoruyko2016wrn} is used for SVHN and STL-10, repectively.
The same code base is adopted as \cite{sohn2020fixmatch} to make the results directly comparable.
Notations follow table \ref{tab:main_cifar}
% $^*$Numbers are generated by \cite{sohn2020fixmatch}. 
% $^\dag$Numbers are produced without CutOut.
% The best number is in bold and the second best number is in italic.
\textbf{Right:}
Error rates on Mini-ImageNet with 40 labels and 100 labels per class.
All methods are evaluated on the same ResNet-18 architecture.
% The best number is in bold and the second best number is in italic.
}
\vspace{-30pt}
\label{tab:main_svhn_stl10_mini}
\end{table*}

% \begin{table}[!]
% \small
% \centering
% % \resizebox{0.76\linewidth}{!}{%
% \begin{tabular}{lrr}
% \toprule
% \multirow{2}{*}{Method} & \multicolumn{2}{c}{Mini-ImageNet} \\ \cmidrule(l){2-3} 
%  & 40 labels & 100 labels \\ \midrule
% Mean Teacher~\cite{tarvainen2017meanteachers} & 72.51$\pm$0.22 & 57.55$\pm$1.11 \\
% Label Propagation~\cite{iscen2019labelpropagation} & 70.29$\pm$0.81 & 57.58$\pm$1.47 \\
% PLCB~\cite{arazo2020plcb} & 56.49$\pm$0.51 & 46.08$\pm$0.11 \\
% FeatMatch~\cite{kuo2020featmatch} & \textit{39.05}$\pm$0.06 & \textit{34.79}$\pm$0.22 \\ \midrule
% CR-Match & \textbf{34.87}$\pm$0.99 & \textbf{32.58}$\pm$1.60 \\ \bottomrule
% \end{tabular}
% % }%
% % \vspace{5pt}
% \caption{Error rates on Mini-ImageNet with 40 labels and 100 labels per class.
% All methods are evaluated on the same ResNet-18 architecture.
% The best number is in bold and the second best number is in italic.
% }
% \label{tab:miniImagenet}
% \end{table}

\vspace{-5pt}
\subsection{Main Results} \label{sec:main_results}
In the following, %This section is organized as follows: 
% \anna{each dataset subsection includes two paragraphs. In each first paragraph we provide technical details, and in each second paragraph we discuss the results.}
each dataset subsection includes two paragraphs.
The first provides technical details and the second discusses  experimental results.
% for each datasets, a description of the datasets and the network architecture are provided in the first paragraph, and the experimental results are discussed in the second paragraph.
% \anna{it should go to implemetnation details}

\myparagraph{CIFAR-10, CIFAR-100, and SVHN.} 
% are widely used SSL datasets with 10, 100, and 10 classes respectively. 
% SVHN is a dataset of 10 digits, which has 73,257 MNIST-like 32-by-32 training images and 531,131 extra and less difficult images. 
% \fan{Since 4, 25 and 100 don't apply to CIFAR-10, I change the text here again:}
We follow prior work \cite{sohn2020fixmatch} and use 4, 25, and 100 labels per class on CIFAR-100 and SVHN without extra data.
For CIFAR-10, we experiment with settings of 4, 25, and 400 labels per class.
% \fan{rewriting ends}
% \bernt{I deleted mentioning of CIFAR-10 - as 4, 25 and 100 applies to various datasets - right? -- ideally you should formulate this more generally like: ``we follow prior with for the various datasets and use 4, 25 and 100 labels per class'' and then cite the respective works}
% We experiments with varying amounts of labeled data from 4 to 100 per class on CIFAR-100 and SVHN without extra data.
% To compare with prior work, we use 4, 25, and 400 labels per class.
We create labeled data by random sampling, and the remaining images are regarded as unlabeled by discarding their labels.
Following~\cite{berthelot2019mixmatch,sohn2020fixmatch,berthelot2019remixmatch}, we use a Wide ResNet-28-2 \cite{zagoruyko2016wrn} with 1.5M parameters on CIFAR-10 and SVHN, and a Wide ResNet-28-8 with 135 filters per layer (26M parameters) on CIFAR-100.

As shown in table \ref{tab:main_cifar} and table \ref{tab:main_svhn_stl10_mini}, our method improves over previous methods across all settings,
% \anna{delete?: of labeled examples and datasets}
and defines a new state-of-the-art.
Most importantly, we improve error rates in low data regimes by a large margin (e.g., with 4 labeled examples per class on CIFAR-100, we outperform FixMatch and the second best method by 9.40\% and 4.83\% in absolute value respectively).
Prior works \cite{sohn2020fixmatch,berthelot2019mixmatch,berthelot2019remixmatch} have reported results using a larger network architecture on CIFAR-100 to obtain better performance.
On the contrary, we additionally evaluate our method on the small network used in CIFAR-10 and find that our method is more than $17$ times ($17\approx26/1.5$) parameter-efficient than FixMatch.
We reach 46.05\% error rate on CIFAR-100 with 4 labels per class using the small model, which is still slightly better than the result of FixMatch using a larger model.

\myparagraph{STL-10.} STL-10 contains 5,000 labeled images of size 96-by-96 from 10 classes and 100,000 unlabeled images.
% , which are extracted from a similar but broader distribution of images, making it more challenging for SSL methods.
The dataset pre-defines ten folds of 1,000 labeled examples from the training data, and we evaluate our method on five of these ten folds as in \cite{sohn2020fixmatch,berthelot2019remixmatch}.
Following~\cite{berthelot2019mixmatch}, we use the same Wide ResNet-37-2 model (comprising 5.9M parameters), and report error rates in table \ref{tab:main_svhn_stl10_mini}. 

Our method achieves state-of-the-art performance with 4.89\% error rate.
Note that FixMatch with error rate 5.17\% used the more advanced CTAugment~\cite{berthelot2019remixmatch}, which learns augmentation policies alongside model training.
When evaluated with the same data augmentation (RandAugment) as we use in CR-Match, our result surpasses FixMatch by 3.09\% (3.09\%=7.98\%-4.89\%), which indicates that CR-Match itself induces a strong regularization effect.
% However, evaluated with the same RandAugment as the strong augmentation, our result surpasses FixMatch by 3.09\% in absolute value (3.09\%=7.98\%-4.89\%). 
% \bernt{whenever differences are discussed I need to pull out a calculator if they are not shown in the table. Much easier is to only point to the absolute numbers in the table} 
% \bernt{I cannot link the text to the table - should be rephrased I guess. In particular I do not see 5,25, and different numbers with different schmes of augmentation fpr CR-Match}

\myparagraph{Mini-ImageNet.}
We follow \cite{iscen2019labelpropagation,arazo2020plcb,kuo2020featmatch} to construct the mini-ImageNet training set.
Specifically, 50,000 training examples and 10,000 test examples are randomly selected for a predefined list of 100 classes~\cite{ravi2016miniImagenet} from ILSVRC~\cite{deng2009imagenet}.
% Each image is center-cropped into resolution 84-by-84. 
% We experiment with settings of 40 labels per class and 100 labels per class by constructing a labeled set and an unlabeled set from the 50,000 training examples as in experiments for CIFAR dataset.
% \bernt{is this setting standard - if yes mention that and give refs} 
Following~\cite{kuo2020featmatch}, we use a ResNet-18 network~\cite{he2016resnet} as our model and experiment with settings of 40 labels per class and 100 labels per class.
% The mean and standard deviation of the error rate is reported in table \ref{tab:miniImagenet}.

% On ImageNet, we follow standard practice~\cite{chen2020simclr, zhai2019s4l} and perform experiments where class-balanced labels are available for only 10 \% of the dataset.

As shown in table \ref{tab:main_svhn_stl10_mini}, our method consistently improves over previous methods and achieves a new state-of-the-art in both the 40-label and 100-label settings.
Especially in the 40-label case, CR-Match achieves an error rate of 34.87\%  which is 4.18\% higher than the second best result.
% Our method surpasses the second best by a large margin. 
% With 40 labeled examples per class, CR-Match achieves 34.89\% which is 4.16\% higher than the second best (39.05\%).
Note that our method is 2 times more data efficient than the second best method FeatMatch \cite{kuo2020featmatch} (FeatMatch, using 100 labels per class, reaches a similar error rate as our method with 40 labeled examples per class).
% \bernt{very short - also no absolute numbers are given even though arguably this is the most important result for many reviewers as this is the most difficult dataset -- please expand}

\vspace{-5pt}
\subsection{Ablation Study} \label{sec:ablation}
\begin{table}
\captionsetup{font=footnotesize,labelfont=footnotesize}
\small
\centering
\vspace{-10pt}
\begin{tabular}{@{}cccccc@{}}
\toprule
\multicolumn{1}{l}{RotNet} & \multicolumn{1}{l}{FeatDistLoss} & MiniImageNet@40 & CIFAR10@4 & CIFAR100@4 & SVHN@4  \\ \cmidrule(l{3pt}r{3pt}){1-2} \cmidrule(l{3pt}r{3pt}){3-3}
\cmidrule(l{3pt}r{3pt}){4-4} \cmidrule(l{3pt}r{3pt}){5-5}
\cmidrule(l{3pt}r{3pt}){6-6}
 &  & 35.13 & 11.86 & 46.22 & 2.42 \\ 
 & \checkmark & 34.14 & \textbf{10.33} & 43.48 & 2.34 \\
\checkmark &  & 34.64 & 11.27 & 41.48 & 2.21 \\
\checkmark & \checkmark & \textbf{33.82}  & 10.92 & \textbf{39.22} & \textbf{2.09}  \\
\bottomrule
\end{tabular}
\vspace{1pt}
\caption{
Ablation studies across different settings.
Error rates are reported for a single split.
}
\vspace{-30pt}
\label{tab:ablation}
\end{table}
In this section, we analyze how FeatDistLoss and RotNet influence the performance across different settings, particularly when there are few labeled samples.
We conduct experiments on a single split on CIFAR-10, CIFAR-100, and SVHN with 4 labeled examples per class, and on MiniImageNet with 40 labels per class.
Specifically, we remove the $\mathcal{L}_{Dist}$ from equation \ref{lu} and train the model again using the same training scheme for each setting.
We do not ablate $\mathcal{L}_{PseudoLabel}$ and $\mathcal{L}_{S}$ due to the fact that removing one of them leads to a divergence of  training.
We present a more detailed analysis of the pseudo-label error rate, the error rate of contributing unlabeled images, and the percentage of contributing unlabeled images during training on CIFAR-100 in the supplementary material.

We report final test error rates in table \ref{tab:ablation}. 
We see that both RotNet and FeatDistLoss contribute to the final performance while their proportions can be different depending on the setting and dataset.
For MiniImageNet, CIFAR-100 and SVHN, the combination of both outperforms the individual losses. For CIFAR-10, FeatDistLoss even outperforms the combination of both. 
%For example, on CIFAR-10, RotNet alone improves the performance, but it does not improve the performance further when FeatDistLoss is used. On CIFAR-100, the error rate is largely reduced by RotNet (4.74\%) while FeatDistLoss also gives another 2.26\% performance boost.
This suggests that RotNet and FeatDistLoss are both important components for CR-Match to achieve the state-of-the-art performance.

\vspace{-5pt}
\subsection{Influence of Feature Distance Loss} \label{sec:analysis}
% \bernt{renamed this section - please check} \fan{Do you mean Influence of Feature Distance \textit{Loss}? Because distance function is only a part of this section.}
In this section, we analyze different design choices for FeatDistLoss to provide additional insights of how it helps generalization.
We focus on a single split with 4 labeled examples from CIFAR-100 and report results for a Wide ResNet-28-2 \cite{zagoruyko2016wrn}. 
For fair comparison, the same 4 random labeled examples for each class are used across all experiments in this section.

\myparagraph{Different distance metrics for FeatDistLoss.}
Here we discuss the effect of different metric functions $d$ for FeatDistLoss.
% We use normalized (cosine similarity) and unnormalized $L2$ distance with respect to maximization and minimization to study equivariance and invariance of the features space respectively.
% \anna{In this section we discuss different approaches to regularize feature space by means of distance function. We use normalized (cosine similarity) and unnormalized $L2$ distance with respect to maximization and minimization to study equivariance and invariance of the features space respectively. }
% \anna{delete?: As is discussed in section \ref{sec:method}, we maximize the cosine distance between features from a weak and a strong augmentation in FeatDistLoss. Here we study the effect of other distance metrics for FeatDistLoss.}
Specifically, we compare two groups of functions in table \ref{tab:analysis_featdistloss} left: metrics that increase the distance between features, including cosine similarity, negative JS divergence, and L2 similarity (i.e. normalized negative L2 distance); metrics that decrease the distance between features, including cosine distance, JS divergence, and L2 distance.
We find that both increasing and decreasing distance between features of different augmentations give reasonable performance. However, increasing the distance always performs better than the counterpart (e.g., cosine similarity is better than cosine distance).
We conjecture that decreasing the feature distance corresponds to an increase of the invariance to data augmentation and leads to ignorance of information like rotation 
% \bernt{is rotation used in data augmentation?}\fan{yes}
or translation of the object. In contrast, increasing the feature distance while still imposing the same label makes the representation equivariant to these augmentations, resulting in more descriptive and expressive representation with respect to augmentation.
Moreover, a classifier has to cover a broader space in the feature space to recognize rather dissimilar images from the same class, which leads to improved generalization.
In summary, we found that both increasing and decreasing feature distance improve over the model which only applies consistency on the classifier level, whereas increasing distances shows better performance by making representations more equivariant.
\begin{table}
\captionsetup{font=footnotesize,labelfont=footnotesize}
\vspace{-10pt}
\resizebox{0.48\textwidth}{!}{%
\begin{tabular}{@{}llr@{}}
        \toprule
        \multicolumn{2}{l}{Metric} & Error rate \\ \midrule
        \multirow{3}{*}{\begin{tabular}[c]{@{}l@{}}Impose \\ equi- \\ variance \end{tabular}} & cosine similarity & \textbf{45.52} \\
         & $L_2$ similarity & 46.22 \\
         & negative JS div. & 46.46 \\
        \midrule
        \multirow{3}{*}{\begin{tabular}[c]{@{}l@{}}Impose \\ invariance\end{tabular}} & cosine distance & 46.98 \\
         & $L_2$ distance & 48.74 \\
         & JS divergence & 47.48 \\ \midrule
        \multicolumn{2}{c}{CR-Match w/o FeatDistLoss} & 48.89 \\ \bottomrule
\end{tabular}
}%
\qquad
\resizebox{0.47\textwidth}{!}{%
\begin{tabular}{@{}lrr@{}}
\toprule
\multirow{2}{*}{Transformations} & \multicolumn{2}{c}{Feature extractor} \\ \cmidrule(l){2-3} 
 & CR-Equiv & CR-Inv \\ \midrule
Translation & 33.22$\pm$0.28 & 36.80$\pm$0.30 \\
Scaling & 11.09$\pm$0.66 & 14.87$\pm$0.40 \\
Rotation & 15.05$\pm$0.33 & 21.92$\pm$0.32 \\
ColorJittering & 31.04$\pm$0.50 & 35.99$\pm$0.27 \\ \bottomrule
\end{tabular}
}%
\vspace{1pt}
\caption{
    \textbf{Left:}
	Effect of different distance functions for FeatDistLoss. The same split on CIFAR-100 with 4 labels per class and a Wide ResNet-28-2 is used for all experiments.
    Metrics that pull features together performs worse than those that push features apart.
    The error rate of CR-Match without FeatDistLoss is shown at the bottom.
    \textbf{Right:}
    Error rates of binary classification (whether a specific augmentation is applied) on the features from CR-Equiv (increasing the cosine distance) and CR-Inv (decreasing the cosine distance).
    We evaluate translation, scaling, rotation, and color jittering. 
    Lower error rate indicates more equivariant features.
    Results are averaged over 10 runs.
    }
\vspace{-30pt}
% \label{tab:pred_aug}
\label{tab:analysis_featdistloss}
\end{table}

\myparagraph{Invariance and equivariance.}
Here we provide an additional analysis to demonstrate that increasing the feature distance provides equivariant features while the other provides invariant features.
Based on the intuition that specific transformations of the input image should be more predictable from equivariant representations,
we quantify the equivariance by how accurate a linear classifier can distinguish between features from augmented and original images.
Specifically, we compare two models from table \ref{tab:analysis_featdistloss} left: the model trained with cosine similarity denoted as \textit{CR-Equiv} % \bernt{I would only introduce the shorthand for both - just CR-Equi - that is good enough and avoids confusion} 
and the model trained with cosine distance denoted as \textit{CR-Inv}. %\bernt{similarly just CR-Inv}
We train a linear SVM to predict whether a certain transformation is applied for the input image.
$1000$ test images from CIFAR-100 are used for training and the rest ($9000$) for validation. 
The binary classifier is trained by an SGD optimizer with an initial learning rate of 0.001 for 50 epochs, and the feature extractor is fixed during training.
We evaluate translation, scaling, rotation, and color jittering in table \ref{tab:analysis_featdistloss} right. 
All augmentations are from the standard PyTorch library. 
The SVM has a better error rate across all augmentations when trained on CR-Equiv features, which means information like object location or orientation is more predictable from CR-Equiv features, suggesting that CR-Equiv produces more equivariant features than CR-Inv.
Furthermore, if the SVM is trained to classify strongly and weakly augmented image features, CR-Equiv achieves a $0.27\%$ test error while CR-Inv is  $46.18\%$.

\myparagraph{Different data augmentations for FeatDistLoss.}
% \fan{table 5 and table 6 will be fine with all comments removed. So please don't change the current layout for table 5 and table 6:)}
In our main results in Section \ref{sec:main_results}, FeatDistLoss is computed between features generated by weak augmentation and strong augmentation.
Here we investigate the impact of FeatDistLoss with respect to different types of data augmentations.
% Here we investigate the question whether it makes more sense to decrease the feature distance if we use the same type of data augmentation.
% If so, should we always use the same kind of augmentation scheme when augmenting the input in our consistency regularization framework?
Specifically, 
we evaluate the error rate of CR-Inv and CR-Equiv under three augmentation strategies:
weak-weak pair indicates that FeatDistLoss uses two weakly augmented images, weak-strong pair indicates that FeatDistLoss uses a weak augmentation and a strong augmentation, and strong-strong pair indicates that
FeatDistLoss uses two strongly augmented images.

%\fan{the following paragraph is rewritten due to the change of the result. Please check.}
% \bernt{I rephrased}
As shown in table \ref{tab:wwss}, using either CR-Inv or CR-Equiv using weak-strong pairs conistently outperforms the other augmentation settings (weak-weak and strong-strong). 
\begin{wraptable}{l}{0.44\textwidth}
\vspace{-12pt}
\captionsetup{font=footnotesize,labelfont=footnotesize}
\begin{tabular}{@{}lrr@{}}
\toprule
Error rate & CR-Inv & CR-Equiv \\ \midrule
% Weak-Weak & 49.14 & 49.98 \\
Weak-Weak & 48.88 & 48.51 \\
Weak-Strong & \textbf{46.98} & \textbf{45.52} \\
% Strong-Strong & 48.82 & 48.82 \\ \bottomrule
Strong-Strong & 48.57 & 48.05 \\ \bottomrule
\end{tabular}
\vspace{-5pt}
\caption{
Effect of combinations of weak and strong augmentation in FeatDistLoss on a Wide ResNet-28-2 for CR-Inv and CR-Equiv. 
}
\vspace{-22pt}
\label{tab:wwss}
% \label{tab:arch_fdl}
\end{wraptable}
Additionally, CR-Equiv consistently achieves better generalization performance across all three settings.
In particular, in the case advocated in this paper, namely using weak-strong pairs, CR-Equiv outperforms CR-Inv by $1.46\%$.
Even in the other two settings, CR-Equiv leads to improved performance even though only by a small margin. 
%When the two versions of the same input do not differ much from each other (e.g. when using weak-weak pairs), CR-Equiv only outperforms CR-Inv by $0.37\%$, however, the gap becomes $1.46\%$ when it comes to weak-strong pairs.
This suggests that, on the one hand, that it is important to use different types of augmentations for our FeatDistLoss. %\bernt{shouldn't we replace `for consistency regularization' here with `for our FeatDistLoss'?}
%Also, enforcing the consistency by decreasing the feature distance appears to be suboptimal when using augmentations of very different types.
And on the other hand, maximizing distances between images that are inherently different while still imposing the same class label makes the model more robust against changes in the feature space and thus gives better generalization performance.

\myparagraph{Linear projection and confidence threshold in FeatDistLoss.}
As mentioned in Section \ref{sec:method}, we apply $\mathcal{L}_{Dist}$ at (a) in Figure~\ref{fig:arch} with a linear layer mapping the feature from the encoder to a low-dimensional space before computing the loss, to alleviate the curse of dimensionality.
Also, the loss only takes effect when the model's prediction has a confidence score above a predefined threshold $\tau$.
Here we study the effect of other design choices in table \ref{tab:arch_fdl}.
% Without the linear projection layer, the error rate increases from 45.52\% to 48.37\%.
% This suggests that a lower dimensional space serves better for comparing distances.
While features after the global average pooling (i.e. (b)) gives a better result than the ones directly from the feature extractor, (b) performs worse than (a) when additional projection heads are added.
Thus, we use features from the feature extractor in CR-Match.
\begin{wraptable}{l}{0.5\textwidth}
\captionsetup{font=footnotesize,labelfont=footnotesize}
\centering
\small
\vspace{-10pt}
\begin{tabular}{@{}crrr@{}}
\toprule
\multicolumn{1}{l}{\begin{tabular}[c]{@{}l@{}}Features taken \\ from Fig.~\ref{fig:arch} at\end{tabular}} & feature & \begin{tabular}[c]{@{}r@{}}feature \\ + linear\end{tabular} & \begin{tabular}[c]{@{}r@{}}feature \\ + MLP\end{tabular} \\
\cmidrule(l{3pt}r{3pt}){1-1} \cmidrule(l{3pt}r{3pt}){2-2} \cmidrule(l{3pt}r{3pt}){3-3} \cmidrule(l{3pt}r{3pt}){4-4}
(a) & 48.37 & \textbf{45.52} & 47.52 \\
(b) & 47.37 & 46.10 & 47.15  \\ \bottomrule
\end{tabular}
% \begin{tabular}{@{}lccccccc@{}}
% \toprule
% \begin{tabular}[c]{@{}l@{}}Features taken \\ from Fig.~\ref{fig:arch} at\end{tabular} & (b) & (b) w/ linear & (b) w/ MLP & (b) w/ linear w/o $\tau$ & (c) & (c) w/ linear & (c) w/ MLP \\ 
% \cmidrule(l{3pt}r{3pt}){1-1} \cmidrule(l{3pt}r{3pt}){2-2} \cmidrule(l{3pt}r{3pt}){3-3} \cmidrule(l{3pt}r{3pt}){4-4}
% \cmidrule(l{3pt}r{3pt}){5-5} \cmidrule(l{3pt}r{3pt}){6-6}
% \cmidrule(l{3pt}r{3pt}){7-7} \cmidrule(l{3pt}r{3pt}){8-8}
% Error rate & 48.37 & \textbf{45.52} & 47.52 & 46.94 & 47.37 & 46.10 & 47.15 \\ \bottomrule
% \end{tabular}
\caption{
Effect of the projection head $z$, and the place to apply $\mathcal{L}_{Dist}$.
(a) denotes un-flattened features taken from the feature extractor directly.
(b) denotes features after the global average pooling.
MLP has 2 FC layers and a ReLU.
Removing the linear projection head harms the test error, and a non-linear projection head does not improve the performance further.
% Effect of the projection head $z$, and the place to apply $\mathcal{L}_{Dist}$.
% (a), (b), and (c) are options where features are taken as shown in Fig.~\ref{fig:arch}.
% (a) is the default choice in CR-match.
% (b) with MLP denotes taking features from (b) and projecting them by a 2-layer MLP before computing $\mathcal{L}_{Dist}$.
% Removing the linear projection head or the confidence threshold harms the test error, and a non-linear projection head or taking features after the global average pooling does not improve the performance further.
}
\vspace{-20pt}
\label{tab:arch_fdl}
\end{wraptable}
The error rate increases from 45.52\% to 48.37\% and 47.52\% when removing the linear layer and replacing the linear layer by a MLP (two fully-connected layers and a ReLU activation function), respectively.
This suggests that a lower dimensional space serves better for comparing distances, but a non-linear mapping does not give further improvement.
% When replacing the linear mapping by a non-linear function which consists of two fully-connected layers and a ReLU activation function \cite{nair2010relu}, the error rate increases to 47.52\%.
% This is probably because a non-linear mapping does not faithfully represent the distance in the original feature space where the final classifier works.
% When using features after the global average pooling layer, the error rate increases to ???\% \fan{needs update}. 
Moreover, when we apply FeatDistLoss for all pairs of input images by removing the confidence threshold, the test error increases from 45.52\% to 46.94\%, which suggests that regularization should be only performed on features that are actually used to update the model parameters, and ignoring those that are also ignored by the model.

\section{Conclusion}
\vspace{-5pt}
The idea of consistency regularization gives rise to many successful works for SSL \cite{bachman2014firstconsistregular,laine2016pimodel,sajjadi2016firstconsistregular,sohn2020fixmatch,xie2019uda,kuo2020featmatch}.
While making the model invariant against input perturbations induced by data augmentation gives improved performance, the scheme tends to be suboptimal when augmentations of different intensities are used.
In this work, we propose a simple yet effective improvement, called FeatDistLoss.
%, to the existing framework.
It introduces consistency regularization on both the classifier level, where the same class label is imposed for versions of the same image, and the feature level, where distances between features from augmentations of different intensities is increased.
By encouraging the representation to distinguish between weakly and strongly augmented images, FeatDistLoss encourages more equivariant representations, leading to improved classification boundaries, and a more robust model.

Through extensive experiments we show the superiority of our training framework, and define a new state-of-the-art on CIFAR-10, CIFAR-100, SVHN, STL-10 and Mini-ImageNet.
Particularly, our method outperforms previous methods in low data regimes by significant margins, e.g., on CIFAR-100 with 4 annotated examples per class, our error rate (39.45\%) is 4.83\% better than the second best (44.28\%).
In future work, we are interested in integrating more prior knowledge and stronger regularization into SSL to further push the performance in low data regimes.
% \fan{the paper length is exactly 12 pages, removing the comments - 20:48 on 23.06.2021}
\bibliographystyle{splncs04}
\bibliography{main}

\end{document}

%%%%%%%%%%%%%%%%%%%%% Add submission id, track, and title. %%%%%%%%%%%%%%%%%%%%%

% Insert your submission number here
\def\SubNumber{060}

% Choose one track by uncommenting one of the following lines  
\def\GCPRTrack{Regular Track}
% \def\GCPRTrack{Track: Computer vision systems and applications}
% \def\GCPRTrack{Track: Pattern recognition in the life and natural sciences}
% \def\GCPRTrack{Track: Photogrammetry and remote sensing}
% \def\GCPRTrack{Track: Robot vision}
% \def\GCPRTrack{Track: DAGM Young Researcher Forum}

% Replace with your title
\title{Appendix: Revisiting Consistency Regularization for Semi-Supervised Learning}
% You can use \thanks for acknowledgment. Do not add any acknowledgment to the draft 
% version that is used for the review process.  
%\title{Title\thanks{XXX}}

\ifreview
	% ANONYMOUS SUBMISSION FOR REVIEW
	% DO NOT MODIFY these for the draft version that is used for the review process.
	\titlerunning{DAGM GCPR 2021 Submission \SubNumber{}. CONFIDENTIAL REVIEW COPY.}
	\authorrunning{DAGM GCPR 2021 Submission \SubNumber{}. CONFIDENTIAL REVIEW COPY.}
	\author{DAGM GCPR 2021 - \GCPRTrack{}}
	\institute{Paper ID \SubNumber}
\else

	\author{First Author\inst{1}\orcidID{0000-1111-2222-3333} \and
	Second Author\inst{2,3}\orcidID{1111-2222-3333-4444} \and
	Third Author\inst{3}\orcidID{2222--3333-4444-5555}}
	
	\authorrunning{F. Author et al.}
	
	\institute{Princeton University, Princeton NJ 08544, USA \and Springer Heidelberg, Tiergartenstr. 17, 69121 Heidelberg, Germany
	\email{lncs@springer.com}\\
	\url{http://www.springer.com/gp/computer-science/lncs} \and ABC Institute, Rupert-Karls-University Heidelberg, Heidelberg, Germany\\
	\email{\{abc,lncs\}@uni-heidelberg.de}}
\fi

\maketitle              % typeset the header of the contribution
\appendix
The supplementary material is organized as follows: 
Section \ref{sec:impl_details} presents more implementation details of our method for reproduciblity.
We report the cosine distance between features from differently augmented images in section \ref{sec:feat_dist}.
Section \ref{sec:tsne} shows t-SNE of input image features extracted by a CR-Match model trained without FeatDistLoss and a CR-Match model with it.
More results of the ablation study of each component of CR-Match are shown in section \ref{sec:ablate_crmatch}.
We present the effect of different confidence thresholds and loss weights in section \ref{sec:loss} and section \ref{sec:tau}, respectively.
We provide pseudo-code in section \ref{sec:pseudo-code}.
Section \ref{sec:aug} shows the list of transformations used for strong augmentation and visualizes examples of strongly and weakly augmented images.

\section{Implementation Details} \label{sec:impl_details}
We minimize the cosine similarity in FeatDistLoss, and use a fully-connected layer for the projection layer $z$, which maps the feature from the original un-flattened 8192-dimension space into a 128-dimension space, the same dimension as the feature dimension for classification.
The predictor head $h$ in rotation prediction loss consists of two fully-connected layers and a ReLU as non-linearity.
We use the same $\lambda_{u}=\lambda_{r}=1$ in all experiments since CR-Match shows good robustness within a range of loss weights in our preliminary experiments.
We train our model for 512 epochs on CIFAR-10, CIFAR-100, and SVHN.
On STL-10 and Mini-ImageNet, we train the model for 300 epochs.
Other hyper-parameters are from \cite{sohn2020fixmatch} for the compatibility.
Specifically, the confidence thresholds $\tau$ for pseudo-label selection is 0.95.
We use SGD with momentum 0.9 and cosine learning rate schedule from \cite{sohn2020fixmatch} starting from 0.03, batch size $B_s$ is 64 for labeled data, and $B_u$ is $7 \times B_s$.
The final performance is reported using an exponential moving average of model parameters as recommended by~\cite{tarvainen2017meanteachers}.

\section{Distance in Feature Space} \label{sec:feat_dist}
In this section, we report the distance between features from differently augmented images for CR-Inv. (increasing the feature distance) and CR-Equiv. (decreasing the feature distance). 
Specifically, we measure the cosine distance between three pairs of features: weak-original pair, strong-original pair, and weak-strong pair.
Features are taken from Fig. 2 from the main paper at (c), which is the space where the classifier works.

Table \ref{tab:cos_dist} summarizes the mean and the standard deviation of the distance from the test set of CIFAR-100.
Compared to CR-Inv., CR-Equiv. enlarges the distance between features from strongly augmented images and weakly augmented (or original), while still ensuring the distance is within a reasonable range (less than 90 degrees). 
This indicates that representations learned from CR-Equiv. are more distinguishable but still close enough to form one cluster.
Also, weak features are always closer to origin features than strong features due to their visual similarities.

\begin{table}[!]
\centering
\begin{tabular}{@{}lrr@{}}
\toprule
Avg. distance & CR-Inv. & CR-Equiv. \\ \midrule
cos($f(\alpha(\textbf{u}))$, $f(\textbf{u})$) & 0.97$\pm$0.01 & 0.96$\pm$0.01 \\
cos($f(\mathcal{A}(\textbf{u}))$, $f(\textbf{u})$) & 0.90$\pm$0.06 & 0.84$\pm$0.06 \\
cos($f(\alpha(\textbf{u}))$, $f(\mathcal{A}(\textbf{u}))$) & 0.90$\pm$0.06 & 0.85$\pm$0.08 \\ 
\bottomrule
\end{tabular}
\vspace{2pt}
\caption{
The cosine distance between features from differently augmented images for CR-Inv. and CR-Equiv.. 
$\textbf{u}$, $\alpha(\textbf{u})$, and $\mathcal{A}(\textbf{u})$ represent the original image, the weakly augmented image, and the strongly augmented images, respectively.
Strong features from CR-Equiv. are further away from weak (or original) features, thus, more distinguishable in the feature space.
}
\vspace{-50pt}
\label{tab:cos_dist}
\end{table}

\section{FeatDistLoss improves decision boundaries.} \label{sec:tsne}
\begin{figure}[!]
	\centering
	\vspace{-10pt}
	\includegraphics[width=\linewidth]{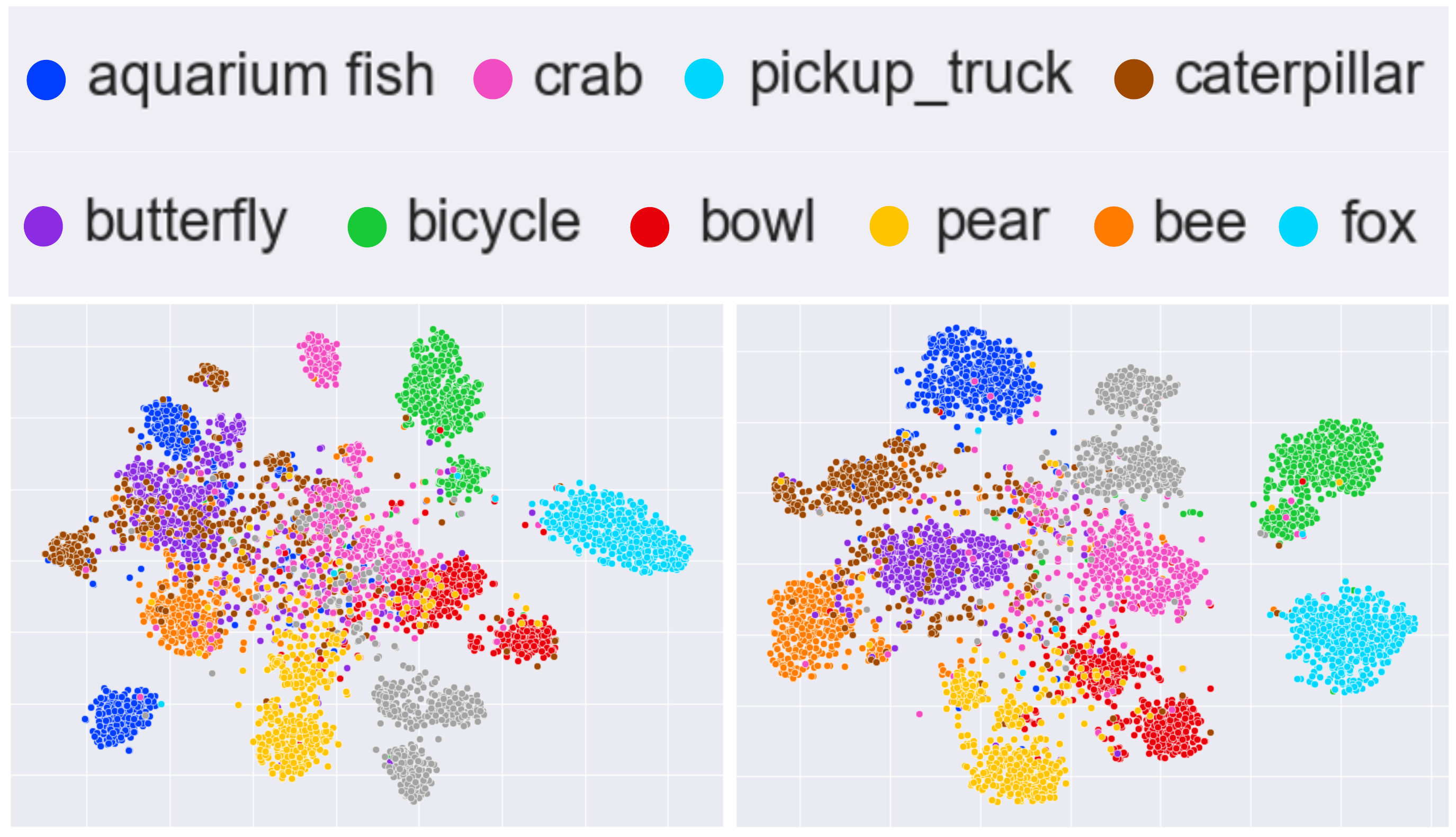}
	\caption{
	We plot t-SNE of input image features extracted by a CR-Match model trained without FeatDistLoss (left) and a CR-Match model with it (right). 
	The better separation from CR-Match suggests that FeatDistLoss improves decision boundaries. 
	}
	\vspace{-10pt}
    \label{fig:tsne}
\end{figure}
% As suggested by figure \ref{fig:teaser}, models trained with FeatDistLoss tend to have improved decision boundaries.
Here we take two models from section 4.2, CR-Match (39.22\% error rate) and CR-Match without FeatDistLoss (41.48\% error rate), and plot t-SNE plots of features extracted from unlabeled images.
As shown in figure~\ref{fig:tsne}, CR-Match with FeatDistLoss produces better separation between classes.
For example, CR-Match forms two clearer clusters for caterpillar and butterfly, while CR-Match without FeatDistLoss mostly mixes them up.
Another example is that the overlap between crab, bowl, and pear is much less for CR-Match compared to CR-Match without FeatDistLoss.
Moreover, the improved decision boundaries also lead to better per-class error rate.
The standard deviation of per-class error rates for CR-Match is 4.34\% lower than that from CR-Match without FeatDistLoss (30.83\% v.s. 26.49\%).

\section{Ablation Study of CR-Match} \label{sec:ablate_crmatch}
\begin{figure*}
\centering
	\includegraphics[width=0.7\linewidth]{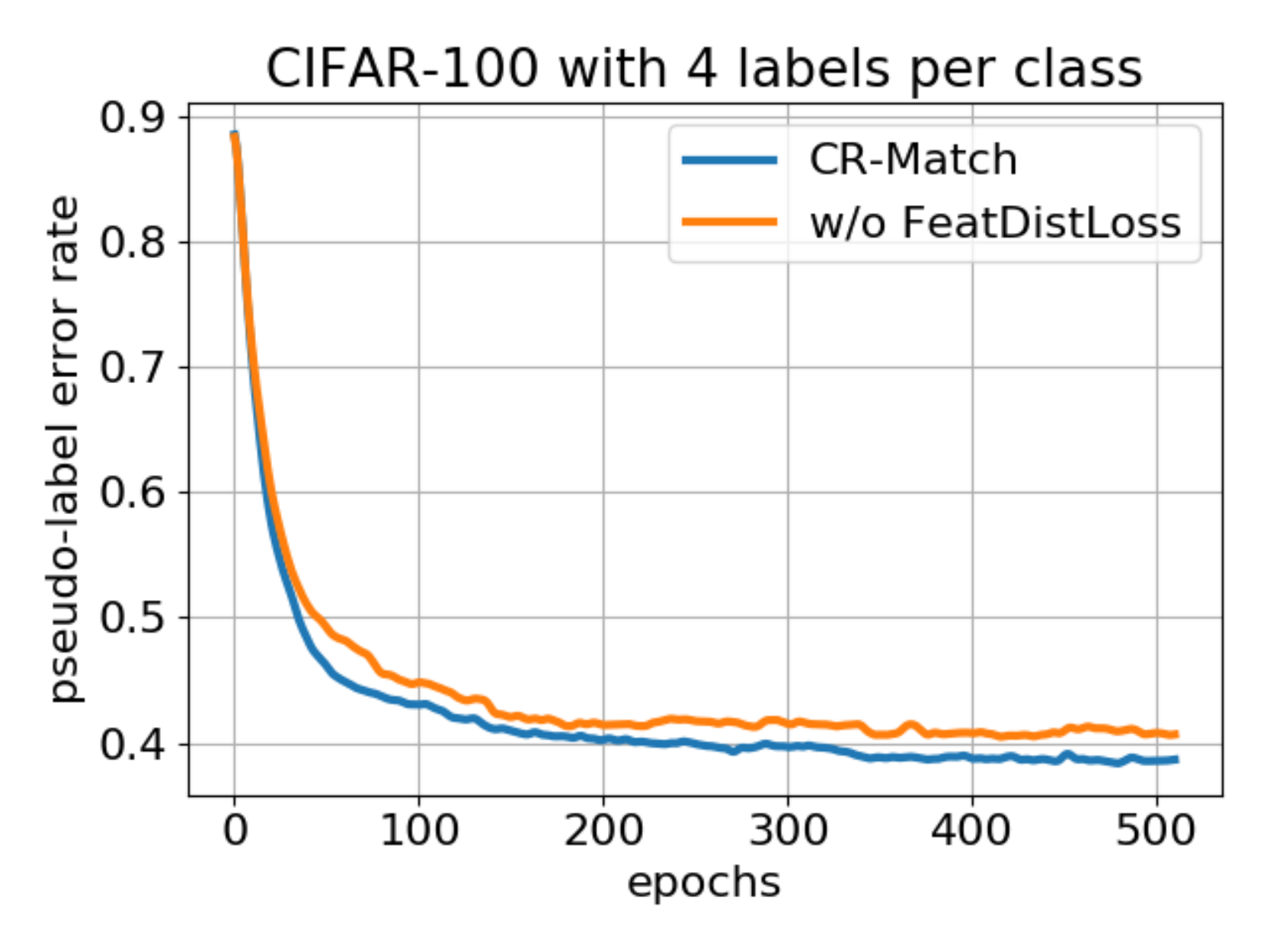}
	\includegraphics[width=0.7\linewidth]{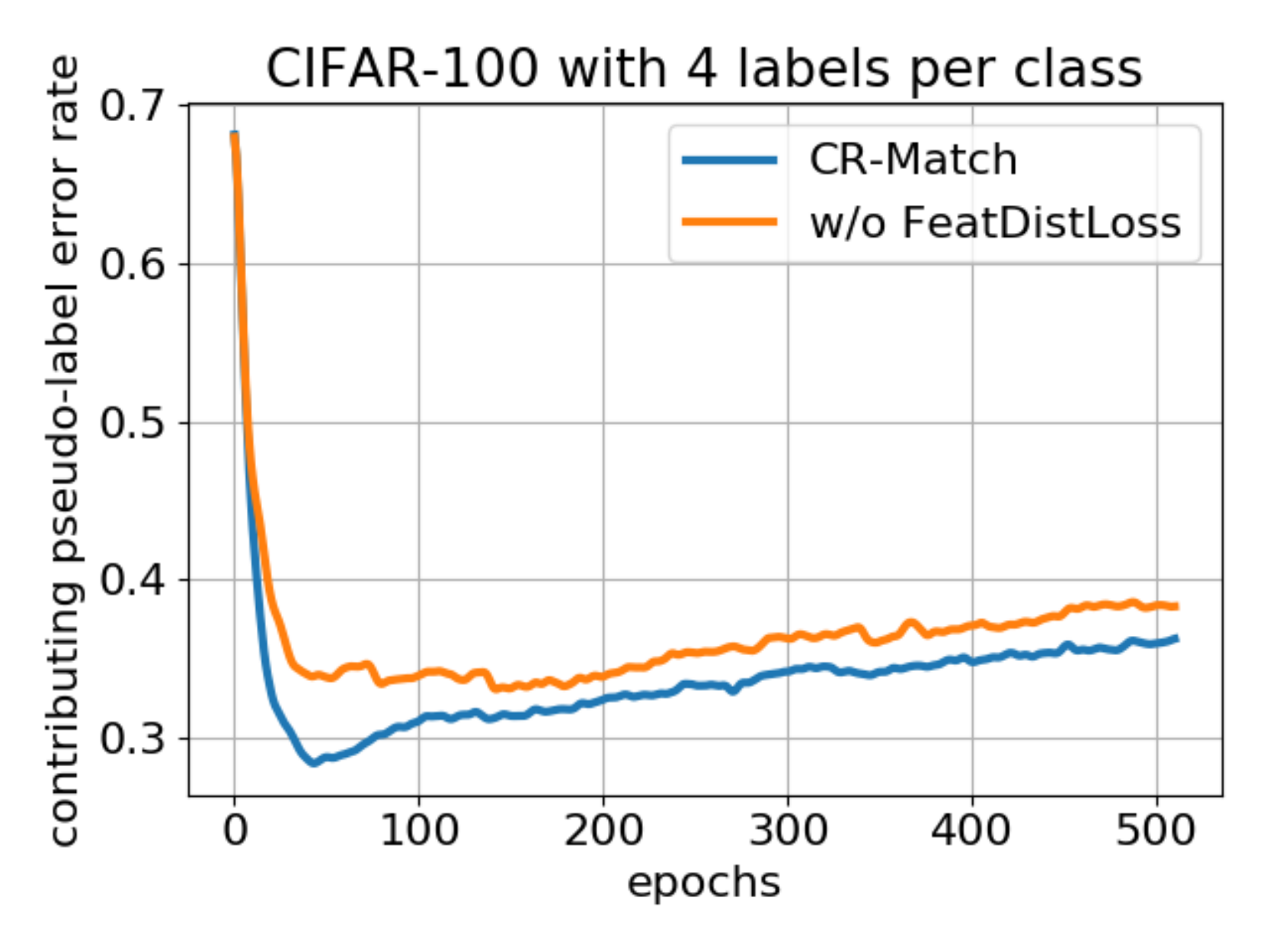}
	\includegraphics[width=0.7\linewidth]{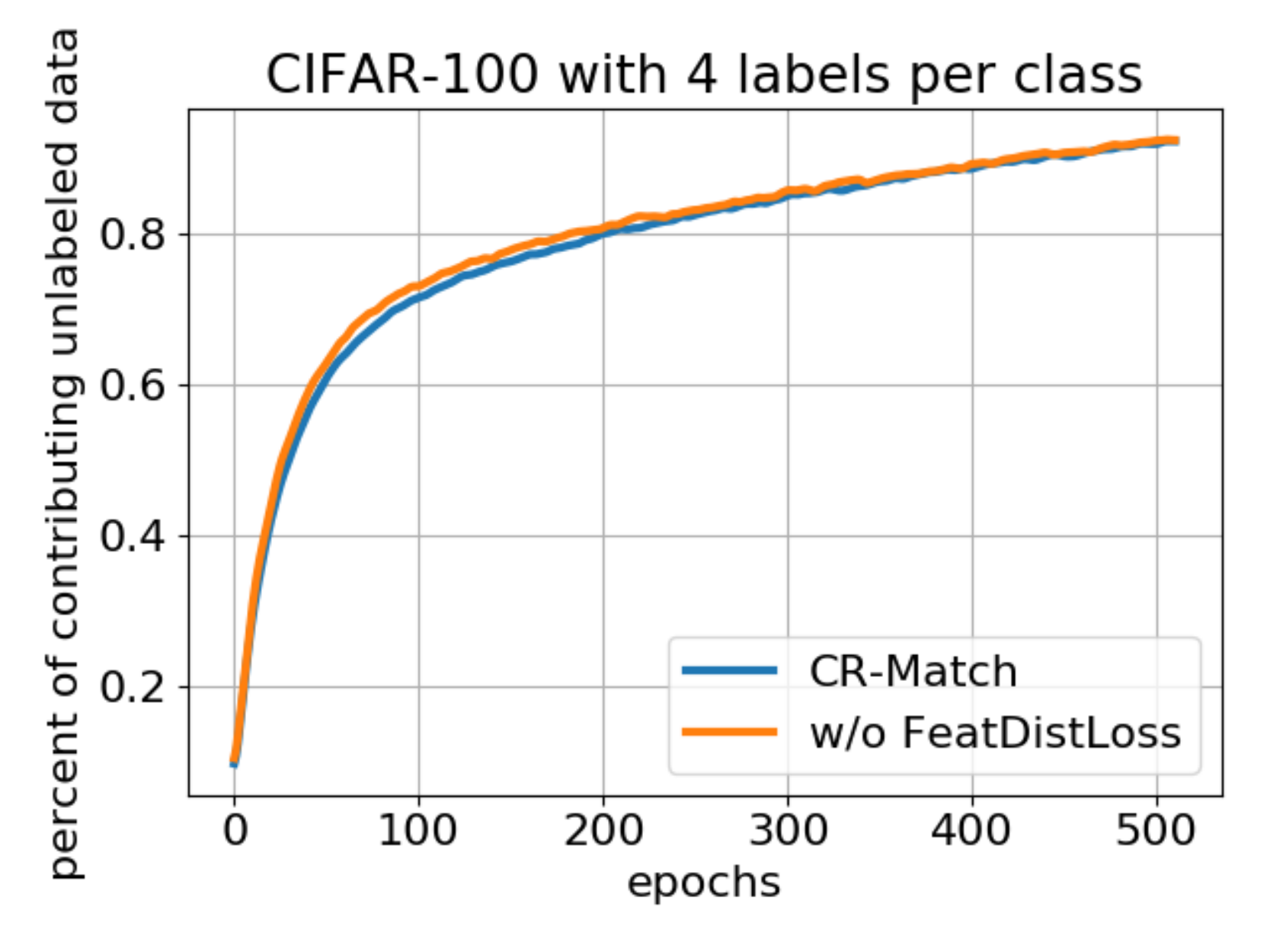}
\caption{Ablation study of our best model on CIFAR-100 with 4 labels per class. CR-Match has a lower pseudo-label error rate as is shown on the top plot. If only the confident predictions are taken into account, CR-Match outperforms the other with a even larger margin in terms of pseudo-label error rate as is shown in the middle. In spite of a better pseudo-label error rate on contributing unlabeled images, the percentage of contributing unlabeled images is maintained the same for CR-Match as is shown on the bottom plot.
}
\label{fig:ablation}
\end{figure*}
As is mentioned in section 4.2, where we ablate the best model on CIFAR-100 with 4 labeled examples per class, % and analyze how different components of CR-Match influence the performance.
we show here a more detailed analysis for CR-Match and CR-Match without FeatDistLoss.

The confidence threshold in CR-Match filters out unconfident predictions during the training.
Therefore, at each training step only images with confidence scores above the threshold contribute to the loss.
% not all images in the batch will contribute to the loss.
We show in figure \ref{fig:ablation} the pseudo-label error rate (top), the error rate of contributing unlabeled images (middle), and the percentage of contributing unlabeled images (bottom) during training.
We observe that CR-Match improves pseudo-labels for the unlabeled data, as it achieves a lower error rate of all unlabeled images and contributing unlabeled images during the training while maintaining the percentage of contributing images.
% While CR-Match does not have the highest percent of contributing unlabeled data, it achieves the lowest error rate of contributing unlabeled images during the training.

\newpage

\section{Effect of Different Confidence Thresholds} \label{sec:tau}
For the main results in the paper, we use a confidence threshold $0.95$ following \cite{sohn2020fixmatch}.
In this section, we study the model robustness against different confidence thresholds.
Experiments are conducted on a single split with 4 labeled examples from CIFAR-100 on a Wide ResNet-28-2.
Figure \ref{tab:lossweights} shows that the error rate of CR-Match when using a confidence threshold from $0.90$ to $0.99$, implying that our model is quite robust against small changes in the confidence threshold. %\bernt{how this is different or similar to which part of the main paper is not mentioned here}  

\begin{figure}
\centering
\includegraphics[width=0.7\linewidth]{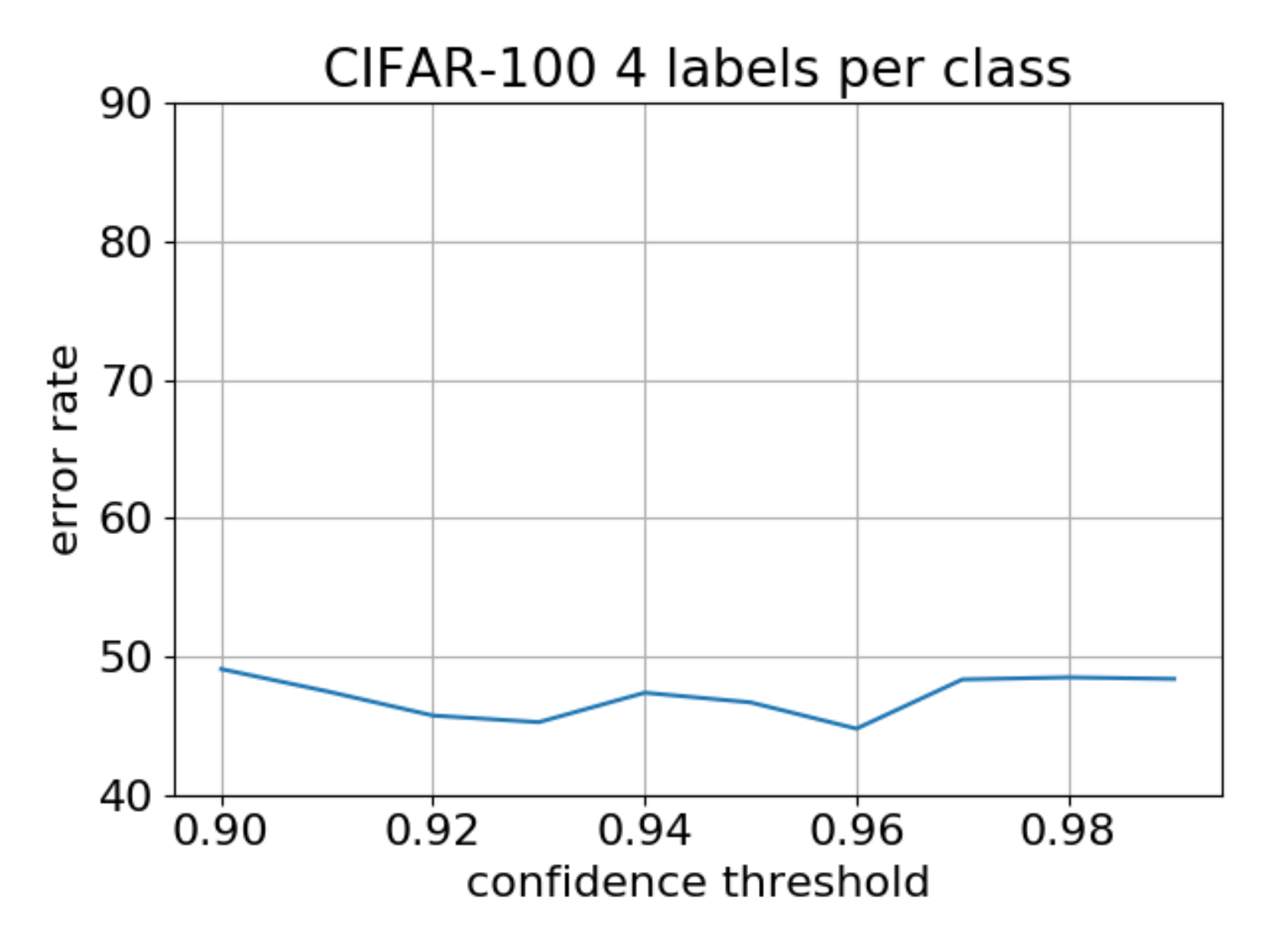}
\caption{
Effect of different confidence thresholds on error rate.
We run experiments on a single split of CIFAR-100 with 4 labels per class.
The model is a Wide-ResNet-28-2.
Our model shows good robustness against small changes in the confidence threshold.
}
\label{fig:tau}
\end{figure}

\section{Effect of Different Loss Weights} \label{sec:loss}
We use $\lambda_{u}=\lambda_{r}=1$ in all experiments in the main paper.
In this section, we test CR-Match with different loss weights.
We report results on a single split with 4 labeled examples from CIFAR-100 on a Wide ResNet-28-2.
CR-Match shows good robustness against different weightings of the rotation prediction loss and the feature distance loss as is shown in table \ref{tab:lossweights}.
The error rate varies from 47.85\% to 45.52\% when weights for the feature distance loss and the rotation prediction loss vary from 5 to 0.1, which shows the feature distance loss is robust against different weighting schemes. %\bernt{how this is different or similar to which part of the main paper is not mentioned here}  

\begin{table}
\centering
\begin{tabular}{@{}ccr@{}}
\toprule
\multicolumn{2}{c}{Loss Weights} & \multirow{2}{*}{Error Rate} \\ \cmidrule(r){1-2}
RotNet & FeatDistLoss &  \\ \midrule
1 & 0.1 & 47.85 \\ 
1 & 5 & 47.66 \\ 
1 & 1 & 45.52 \\
0.1 & 1 & 46.73 \\ 
% 5 & 1 & 56.91 \\ 
% 10 & 1 & 60.15  \\ 
% 1 & 10 & 44.71 \\
% 0.5 & 1 & 46.73 \\ 
% 1 & 0.5 & 46.68 \\ 
\bottomrule
\end{tabular}
\vspace{2pt}
\caption{
Effect of different loss weights for the rotation prediction loss and the feature distance loss.
Our model shows good robustness against different weighting schemes.
We run experiments on a single split of CIFAR-100 with 4 labels per class.
The model is a Wide-ResNet-28-2.
}
\label{tab:lossweights}
\end{table}

\section{Pseudo-Code} \label{sec:pseudo-code}
As is promised in section 3.2, we present here the complete algorithm of CR-Match processing one batch of labeled and unlabeled images in algorithm \ref{alg}.

\begin{algorithm*}[ht]
\caption{}
    \begin{algorithmic}[1]  \label{alg}
    % \footnotesize
    \STATE {\bfseries Input:} Labeled batch $\mathcal{X} = \big\{(\textbf{x}_i, \textbf{p}_i): i \in (1,\ldots,B_s)\big\}$, unlabeled batch $\mathcal{U} = \big\{\textbf{u}_i: i \in (1,\ldots,B_u)\big\}$, confidence threshold $\tau$, FeatDistLoss weight $\lambda_{u}$, rotation prediction loss weight $\lambda_{r}$, classifier $g$, distance metric $d$, FeatDistLoss head $z$, rotation prediction head $h$. \\
    \STATE $\mathcal{L}_S = \frac{1}{B_s} \sum_{i=1}^{B_s} \ell_{CE}(\textbf{p}_i, g(\alpha(\textbf{x}_i)))$ \COMMENT{\textit{Cross-entropy loss for labeled data}} \\
    \FOR{$i = 1$ \TO $B_u$}
    \STATE $\textbf{u}_i^w=f(\alpha(\textbf{u}_i))$ \COMMENT{\textit{Extract representation from weak data augmentation}} \\
    \STATE $\textbf{u}_i^s=f(\mathcal{A}(\textbf{u}_i))$ \COMMENT{\textit{Extract representation from strong data augmentation}} \\
    \STATE $c_i=max \ g(\textbf{u}_i^w)$ \COMMENT{\textit{Compute confidence score from the weakly augmented image}} \\
    \ENDFOR
    \STATE $\mathcal{L}_{PseudoLabel} = \frac{1}{B_u} \sum_{i=1}^{B_u} \mathbbm{1}\{c_i>\tau\}\  \ell_{CE}(g(\textbf{u}_i^w), \textbf{u}_i^s)$ \COMMENT{\textit{Cross-entropy loss with pseudo-label for unlabeled data}}
    \STATE $\mathcal{L}_{Dist} = \frac{1}{B_u} \sum_{i=1}^{B_u} \mathbbm{1}\{c_i>\tau\}\  d(z(\textbf{u}_i^w),z(\textbf{u}_i^s) )$ \COMMENT{\textit{Decrease or increase the feature distance for unlabeled data}}
    \STATE $\mathcal{L}_{Rot} = \frac{1}{4B_u} \sum_{i=1}^{B_u} \sum_{r \in \mathbbm{R}} \ell_{CE}(Rotate(\textbf{u}_i^w,r), h(\textbf{u}_i^w))$ \COMMENT{rotation prediction loss}
\RETURN $\mathcal{L}_S + \lambda_{u} (\mathcal{L}_{PseudoLabel} + \mathcal{L}_{Dist}) + \lambda_{r} \mathcal{L}_{Rot}$
\end{algorithmic}
\end{algorithm*}

\section{List of Data Transformations and Examples of Augmented Images} \label{sec:aug}
\begin{figure*}[h]
    \centering
    \includegraphics[width=1\linewidth]{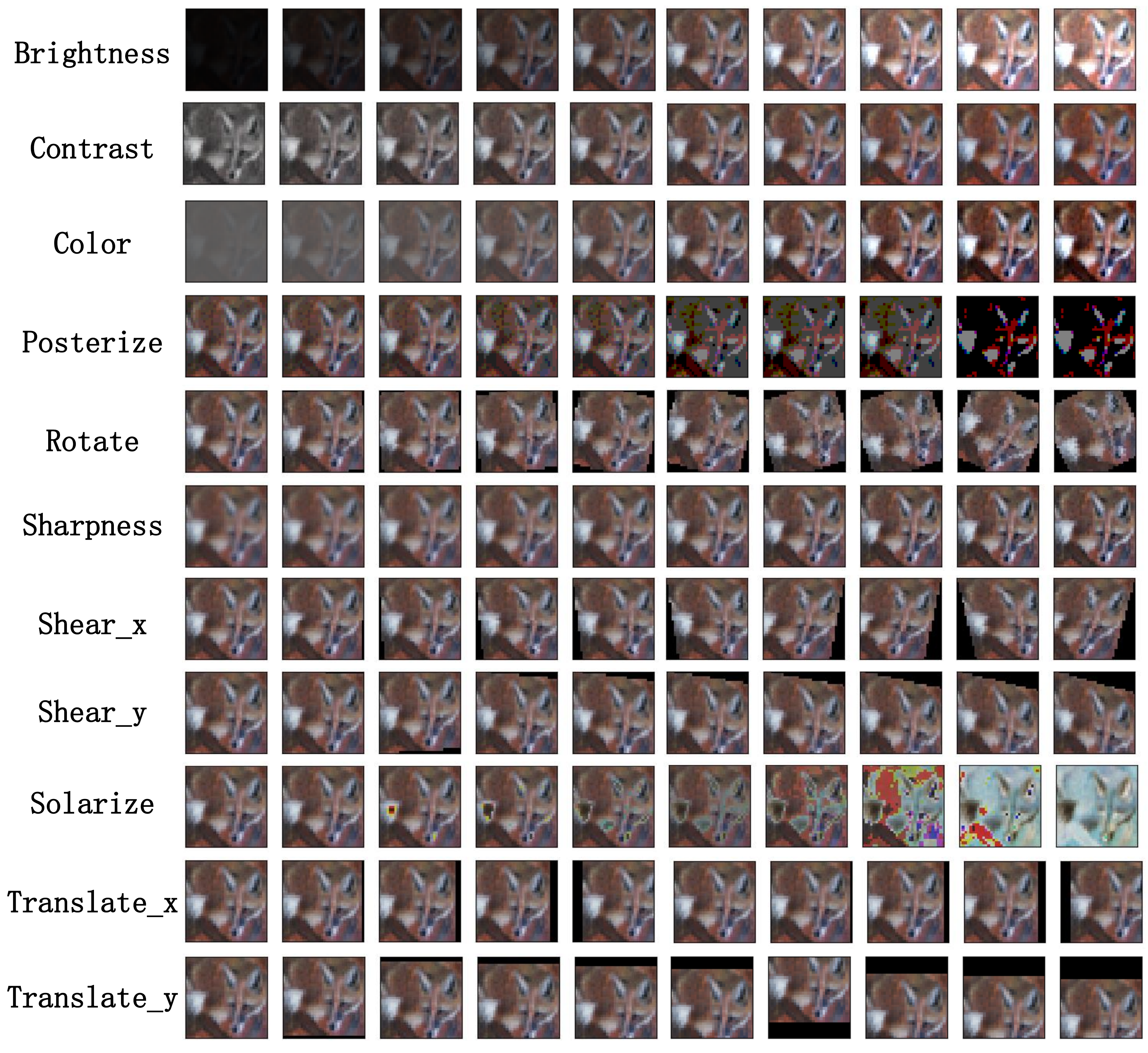}
    \caption{Visualization of effects of different transformations with different magnitudes used in strong augmentation. The original image is taken from CIFAR-100.}
    \label{fig:augment}
\end{figure*}

As mentioned in section 3.3, we list all transformation operations with corresponding ranges used for strong data augmentation in table \ref{tab:randaugment_ops} and visualize effects of different transformations with different magnitudes on an image from CIFAR-100 in figure \ref{fig:augment}.
We used the same sets of image transformations used in \cite{sohn2020fixmatch}.

\begin{figure*}[h]
    \centering
    \includegraphics[width=1\linewidth]{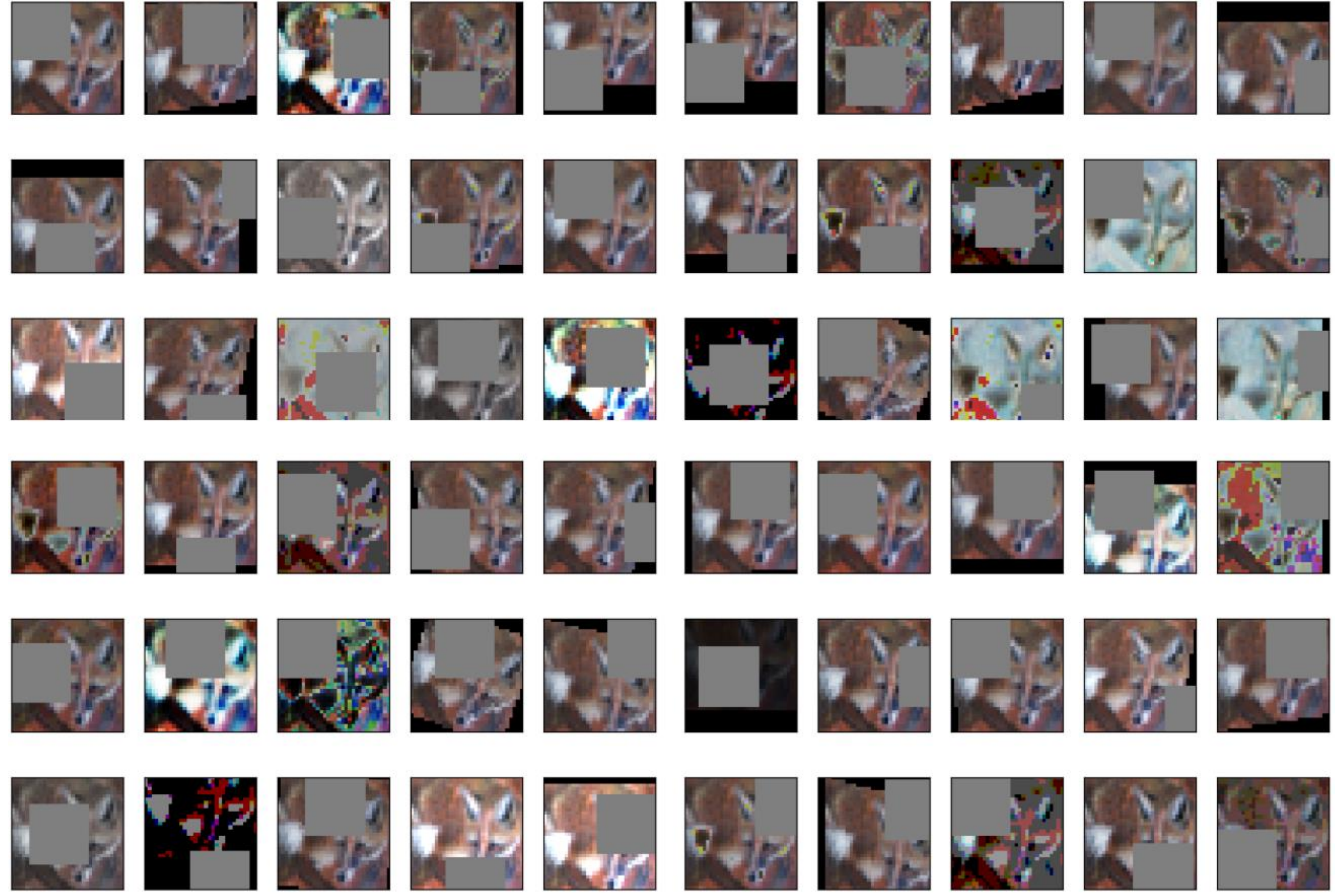}
    \caption{Examples of strongly augmented images. The original image is taken from CIFAR-100.}
    \label{fig:strong}
\end{figure*}

\begin{figure*}[h]
    \centering
    \includegraphics[width=1\linewidth]{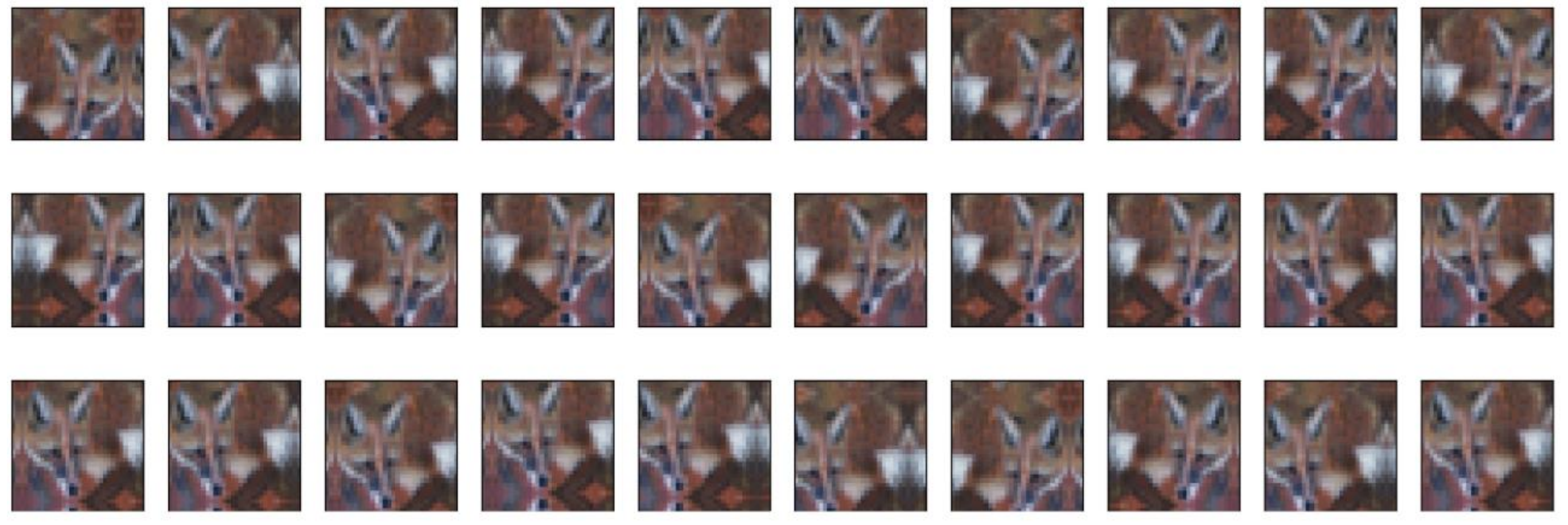}
    \caption{Examples of weakly augmented images. The original image is taken from CIFAR-100.}
    \label{fig:weak}
\end{figure*}

Figure \ref{fig:strong} and figure \ref{fig:weak} show 60 randomly picked examples from strong augmentation and weak augmentation, respectively.
% Examples of strongly augmented images from the weak augmentation and the strong augmentation are show in figure \ref{fig:augment}.
The original image is from CIFAR-100.

\begin{table*}[ht]
\small
\footnotesize
\resizebox{\textwidth}{!}{%
\begin{tabular}{lp{10cm}lp{2cm}}
\toprule
Transformation             & Description    & Parameter &         Range \\
\midrule
Autocontrast &Maximizes the image contrast by setting the darkest (lightest) pixel to black (white).&        &                              \\
Brightness   &Adjusts the brightness of the image. $B=0$ returns a black image, $B=1$ returns the original image.&$B$&[0.05, 0.95]\\
Color        &Adjusts the color balance of the image like in a TV. $C=0$ returns a black \& white image, $C=1$ returns the original image.&$C$&[0.05, 0.95]                \\
Contrast     &Controls the contrast of the image. A $C=0$ returns a gray image, $C=1$ returns the original image.& $C$&[0.05, 0.95] \\
Equalize     &Equalizes the image histogram.&&\\
Identity     &Returns the original image. &           &                                  \\
Posterize    &Reduces each pixel to $B$ bits. &$B$&[4, 8] \\
Rotate       &Rotates the image by $\theta$ degrees.& $\theta$    & [-30, 30]      \\
Sharpness    &Adjusts the sharpness of the image, where $S=0$ returns a blurred image, and $S=1$ returns the original image.&$S$&[0.05, 0.95]\\
Shear\_x     &Shears the image along the horizontal axis with rate $R$.& $R$        & [-0.3, 0.3]     \\
Shear\_y     &Shears the image along the vertical axis with rate $R$.& $R$         & [-0.3, 0.3]     \\
Solarize     &Inverts all pixels above a threshold value of $T$.        &$T$           & [0, 1]    \\
Translate\_x &Translates the image horizontally by ($\lambda\times$image width) pixels.  &    $\lambda$&[-0.3, 0.3]\\
Translate\_y &Translates the image vertically by ($\lambda\times$image height) pixels.  &    $\lambda$&[-0.3, 0.3]\\
\bottomrule
\end{tabular}
}%
\caption{List of transformations used in strong data augmentation for CR-Match.}
\label{tab:randaugment_ops}
\end{table*}

\bibliographystyle{splncs04}
\bibliography{egbib}